\documentclass{ieeetj}
\usepackage{cite}
\usepackage{amsmath,amssymb,amsfonts}
\usepackage{graphicx,color}
\usepackage{xcolor}
\usepackage{hyperref}
\usepackage{siunitx}
\hypersetup{hidelinks=true}
\usepackage{array}
\usepackage{textcomp}
\usepackage{stfloats}
\usepackage{url}
\usepackage{verbatim}
\usepackage{multirow}
\usepackage{bm}
\usepackage{multicol}
\usepackage{ragged2e}
\usepackage{booktabs}
\usepackage{lscape}

\usepackage[vlined,noend,ruled]{algorithm2e}
\usepackage{algpseudocode}  
\newcolumntype{P}[1]{>{\justifying\arraybackslash}m{#1}}
\usepackage[compact]{titlesec}  
\newcolumntype{J}[1]{>{\justifying\arraybackslash}p{#1}}

\def\BibTeX{{\rm B\kern-.05em{\sc i\kern-.025em b}\kern-.08em
    T\kern-.1667em\lower.7ex\hbox{E}\kern-.125emX}}
\AtBeginDocument{\definecolor{tmlcncolor}{cmyk}{0.93,0.59,0.15,0.02}\definecolor{NavyBlue}{RGB}{0,86,125}}

\def\OJlogo{\vspace{-4pt}$<$Society logo(s) and publication title will appear here.$>$}
\def\seclogo{\vspace{10pt}$<$Society logo(s) and publication title will appear here.$>$}

\DeclareSIUnit\pixel{px}

\begin{document}
\receiveddate{XX Month, XXXX}
\reviseddate{XX Month, XXXX}
\accepteddate{XX Month, XXXX}
\publisheddate{XX Month, XXXX}
\currentdate{XX Month, XXXX}
\doiinfo{XXXX.2022.1234567}

\title{xFLIE: Leveraging Actionable Hierarchical Scene Representation for Autonomous Semantic-Aware Inspection Missions}

\author{Vignesh Kottayam Viswanathan, Student, IEEE, Mario Alberto Valdes Saucedo , Student, IEEE, \\ Sumeet Gajanan Satpute, Member, IEEE, Christoforos Kanellakis, Member, IEEE and George Nikolakopoulos, Member, IEEE}
\affil{Robotics and AI, Luleå University of Technology 97187, Luleå, Sweden}
\corresp{Corresponding author: Vignesh Kottayam Viswanathan (email: vigkot@ltu.se).}
\authornote{The project is partially funded by the European Union's Horizon Europe Research and Innovation Programme, under the Grant Agreement No. 101091462 M4mining and partially under the Grant agreement No. 101138451 PERSEPHONE}
\begin{abstract}
We present a novel architecture aimed towards incremental construction and exploitation of a hierarchical 3D scene graph representation during semantic-aware inspection missions. Inspection planning, particularly of distributed targets in previously unseen environments, presents an opportunity to exploit the semantic structure of the scene during reasoning, navigation and scene understanding. Motivated by this, we propose the 3D Layered Semantic Graph (3DLSG), a hierarchical inspection scene graph constructed in an incremental manner and organized into abstraction layers that support planning demands in real-time. To address the task of semantic-aware inspection, a mission framework, termed as Enhanced First-Look Inspect Explore (xFLIE), that tightly couples the 3DLSG with an inspection planner is proposed. We assess the performance through simulations and experimental trials, evaluating target-selection, path-planning and semantic navigation tasks over the 3DLSG model. The scenarios presented are diverse, ranging from city-scale distributed to solitary infrastructure targets in simulated worlds and subsequent outdoor and subterranean environment deployments onboard a quadrupedal robot. The proposed method successfully demonstrates incremental construction and planning over the 3DLSG representation to meet the objectives of the missions. Furthermore, the framework demonstrates successful semantic navigation tasks over the structured interface at the end of the inspection missions. Finally, we report multiple orders of magnitude reduction in path-planning time compared to conventional volumetric-map-based methods over various environment scale, demonstrating the planning efficiency and scalability of the proposed approach.
\end{abstract}


\maketitle

\section{Introduction}

\IEEEPARstart{P}{resent} visual inspection methods utilize autonomous robots as data-collection platforms for routine tasks such as geometric reconstruction~\cite{schmid2020efficient,feng2023predrecon}, change detection~\cite{vieira2014spatial}, and site monitoring~\cite{ginting2024seek}. Human operators process the collected information to analyse, summarize, and share updated objectives with the robot (e.g., \textit{observation of region {x,y,z}}). Since humans naturally interpret environments through semantic concepts (e.g., “a crack was observed on the south-facing building facade”), it becomes intuitive to correlate inspection results with real-world entities and plan subsequent actions (e.g., “dispatch a maintenance team to repair it”). Currently, there is a substantial interest within the research community to utilize such semantic concepts to inform robot planning during a task.

We envision the next generation of autonomous inspection system to build and exploit an interpretable scene representation to support advanced planning and reasoning tasks~\cite{ravichandran2024spine}. A promising research avenue towards enabling such scene interpretability is the use of 3D Scene Graphs (3DSGs): structured, hierarchical representations that unify the metric, semantic, and relational properties of a scene~\cite{armeni20193d,rosinol20203d,hughes2023foundations}. Existing 3DSG systems have demonstrated scene graphs as compact yet descriptive
environment representations, recognizing its potential to support efficient planning in large
scenes~\cite{rosinol20203d,Hughes2022HydraAR,agia2022taskography}.
\begin{figure*}[htbp]
    \centering
    \includegraphics[width=\linewidth]{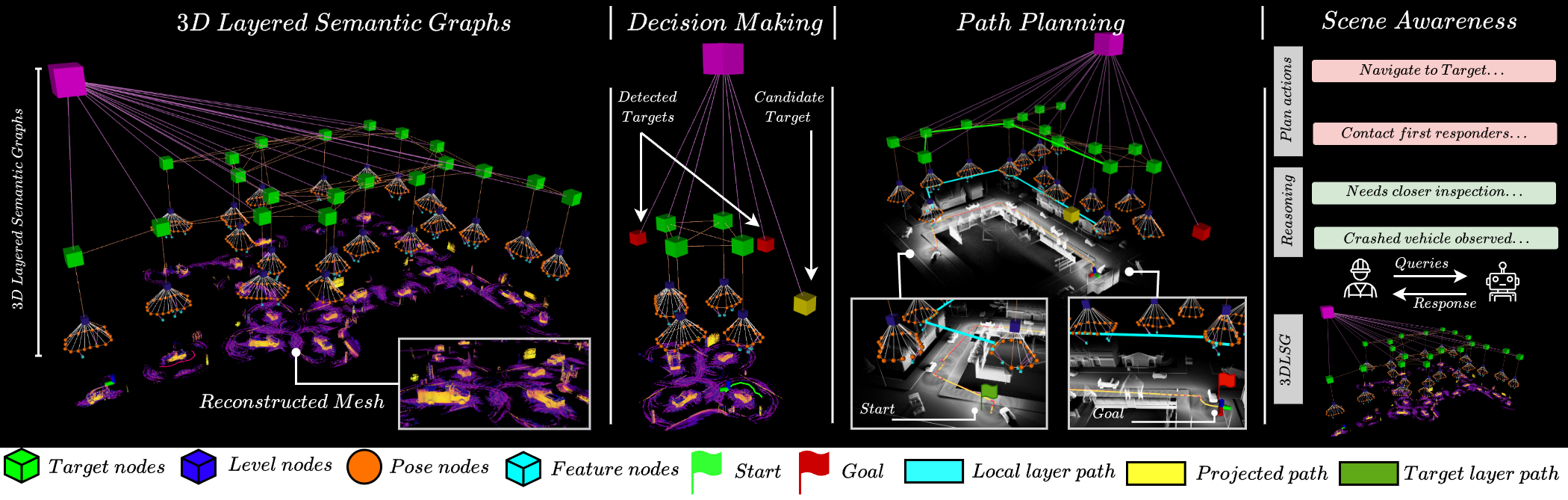}
    \caption{An overview of the inspection scene graph constructed during the semantic-aware inspection mission within a simulated city-scale environment. The middle subfigures illustrate the planning avenues explored during the mission. This includes decision-making and hierarchical path planning over the incremental scene graph representation during autonomous inspection. The rightmost subfigure highlights semantic navigation tasks based on a query from a human operator using the structured scene graph.}
    \label{fig:xflie_overview}
\end{figure*}

However, on examining the progress of planning over 3DSGs, most approaches either assume a complete and a priori known scene graph available for planning~\cite{agia2022taskography,ray2024task,ginting2024seek} or (ii) automate the
construction of the complete 3DSG model prior to planning tasks~\cite{Hughes2022HydraAR,maggio2024clio,gu2024conceptgraphs}. Although some studies jointly construct and plan over a scene graph, they do so in a limited fashion by operating over a sparse graph model~\cite{chen2023not}. These methods recognize
the benefits of scene abstraction for robot planning including decision-making, semantic navigation and
human-robot interaction tasks. Nevertheless, they either confine planning to static, fully known world models or restrict it to sparsely defined semantic structures.

This exposes a gap in existing capability of such semantic scene graph representations to support autonomous missions in unknown environments. Specifically, where a robot usually has partial observation of the scene in the initial phase and is required to concurrently build and plan over the incremental scene model to meet the mission objectives. Motivated by the above, we present Enhanced First-Look Inspect-Explore framework (xFLIE), that couples the incremental construction hierarchical scene graph, termed as 3D Layered Semantic Graph (3DLSG), with an existing inspection planner~\cite{viswanathan2023towards}, First-Look Inspect-Explore (FLIE), to address inspection of distributed semantic targets in unknown environments. Specifically, we define a schema that formalizes the composition and the functionality of these layers. Next, we present a pipeline for incrementally constructing and exploiting the scene model to guide target selection and path planning during inspection of distributed targets. Finally, a quantitative study demonstrating the significant gain in planning efficiency by exploiting such representation for autonomous missions in large-scale environments is presented. Figure~\ref{fig:xflie_overview} presents an overview of the proposed framework, which integrates incremental 3D scene graph construction with an inspection planner to enable semantic-aware inspection in unknown environments.

The proposed framework is evaluated in extensive simulation environments and validated through representative field deployments spanning outdoor urban and subterranean settings. We evaluate the framework for solitary infrastructure inspection (collapsed buildings) as well as within a city scale scene with distributed targets (cars, trucks). In simulation, we benchmark 3DLSG-enabled planning against a conventional voxel-level approach. We analyze planning latency across multiple map resolutions and environment scales. The results demonstrate that the hierarchical abstraction of 3DLSG substantially improves computational efficiency, achieving up to multiple orders of magnitude reduction in planning time compared to voxel-level representations.

In field trials, the proposed architecture is deployed onboard a quadrupedal robot to inspect distributed semantic targets in unstructured outdoor and subterranean environments. During each mission, the robot incrementally constructed the multiple layers of the 3DLSG while successfully performing navigation and target selection tasks over the evolving scene model during inspection. The robot successfully leverages the hierarchical representation to plan paths toward inspection targets in real time and to interpret semantic navigation commands issued by a human operator, demonstrating the feasibility of integrated scene-graph-based planning in real-world missions.

The remainder of this article is organized as follows. Section~\ref{sec:rel_works} outlines the related works, followed by a formal problem statement in Section~\ref{sec:problem}. Section~\ref{sec:3D_LSG} defines and describes the composition of the abstraction layers comprising the 3DLSG. Section~\ref{sec:xFLIE} discusses the integration of 3DLSG with FLIE and explains how each abstraction layer is populated and used during the inspection mission. Section~\ref{sec:eval_setup} presents the evaluation setup used for validation. Section~\ref{sec:res_des} discusses the results obtained, while Section~\ref{sec:lessons} highlights the insights gained. Finally, Section~\ref{sec:future_works} suggests potential improvements, and Section~\ref{sec:conclusions} summarizes the work.
\section{Related Works}\label{sec:rel_works}
Early research on hierarchical scene abstraction for mobile robots established the foundation for representing both spatial and semantic information. \cite{galindo2005multi} introduced a spatial–semantic hierarchical model that organized the environment into two interconnected hierarchies: one capturing spatial structure and the other encoding semantic context. Similarly, \cite{zender2008conceptual} developed a multi-layered semantic map built online from sensor measurements, enabling robots to address natural language queries through semantic grounding.

More recent work has focused on 3D Scene Graphs (3DSGs) as unified environment representations integrating geometric and semantic layers. \cite{armeni20193d} proposed an offline approach for 3D scene graph generation from RGB and geometric data, primarily targeting indoor environments. \cite{rosinol2021kimera} advanced this direction, generating dynamic 3D scene graphs from metric-semantic meshes though higher-level abstractions were obtained via post-processing. Extending \textit{Kimera} to real-time operation, \cite{Hughes2022HydraAR,hughes2023foundations} introduced \textit{Hydra}, achieving online 3DSG construction onboard mobile platforms. Parallel research explored alternative hierarchical representations such as multi-layer visibility graphs~\cite{zhou2022hivg} and factor-graph-based models for improved localization and mapping~\cite{bavle2022situational}. Recent developments have expanded 3DSG capability through open-vocabulary reasoning. \cite{gu2024conceptgraphs} introduced an open-vocabulary 3D scene graph framework enabling downstream planning from posed RGB-D data. \cite{Werby-RSS-24} further extended this paradigm by constructing actionable scene graphs from segmented global point clouds. Despite these advancements, most existing frameworks emphasize perception and mapping, few address closed-loop coupling between scene graph construction and autonomous decision-making during missions.

Efforts to exploit 3DSGs for planning have primarily focused on reasoning over static world models. \cite{agia2022taskography} formalized symbolic reasoning tasks over known 3DSGs, while \cite{ray2024task} introduced a hierarchical planner that extracts a feasible planning domain from a global scene graph to solve multi-step, high-level tasks. \cite{ginting2024seek} investigated probabilistic semantic planning over a pre-built 3D spatial scene graph coupled with relational networks. Other approaches employ scene graphs as contextual priors for planning and search. \cite{amiri2022reasoning} integrated local scene graphs with a Markov network to improve target search, while \cite{chen2023not} used spatial scene graphs for object-goal navigation over a 2D grid map. \cite{gu2024conceptgraphs} demonstrated broad downstream applications including object search and re-localization over global scene graph models. Although these studies validate the utility of structured abstraction for reasoning and navigation, they typically rely on complete and static world representations, limiting their applicability in initially unknown environments.

Planning in unknown environments has been extensively studied within the Next-Best-View (NBV) and coverage planning frameworks within the context of exploration and inspection missions. Information-theoretic sampling approaches aim to maximize the expected information gain per observation while minimizing travel cost and energy consumption~\cite{song2018surface,bircher2018receding,schmid2020efficient,naazare2022online}. These strategies effectively achieve geometric coverage of the environment. Recent advances have introduced semantic awareness into the planning process.~\cite{de2021real} combined Rapidly-exploring Random Tree (RRT) exploration with semantic heuristics to bias viewpoint selection.~\cite{dharmadhikari2023semantics} developed a frontier-based volumetric exploration algorithm integrated with semantics-aware coverage planning, while \cite{ginting2024semantic} introduced a belief–behaviour graph framework for inspecting semantic targets under observation uncertainty. Other studies emphasize object-centric or semantic-risk-aware mapping. \cite{papatheodorou2023finding} employed a weighted NBV strategy to balance volumetric exploration and semantic mapping for 3D reconstruction of target objects. \cite{lu2024semantics} proposed a multi-layer, object-centric volumetric mapping pipeline that extracts semantic frontiers, and \cite{hu2025semantic} constructed an offline metric–semantic voxel map later used by a semantic-risk-aware A* planner for language-based inspection goals. Beyond object mapping, several works demonstrate semantic localization over volumetric maps for contextual inspection missions~\cite{yang2018semantic,yang2023automated,ge2025deep}.

In summary, current 3DSG systems have shown that scene graphs provide compact yet descriptive representations of 3D environments, highlighting their potential to support efficient planning in large-scale scenes~\cite{rosinol20203d,Hughes2022HydraAR,agia2022taskography}. However, existing approaches either (i) assume a complete and a priori known scene graph for planning~\cite{agia2022taskography,ray2024task,ginting2024seek}, or (ii) construct a static world model prior to executing planning tasks~\cite{Hughes2022HydraAR,maggio2024clio,gu2024conceptgraphs}. On the other hand, although planning methods demonstrate the benefits of incorporating semantic context into inspection~\cite{yang2018semantic,yang2023automated,ge2025deep} and navigation~\cite{chen2023not,dharmadhikari2023semantics,ginting2024semantic}, they directly rely on occupancy maps for either navigation or for scene representation and do not fully exploit the efficient planning foundation that 3DSGs can provide. Furthermore, current inspection methods tend to focus mainly on producing a geometric model of the scene captured during the mission, treating this as the primary goal~\cite{papatheodorou2023finding,dharmadhikari2023semantics}. Consequently, semantic observations are used primarily to guide the planning process toward achieving this objective. In contrast, we argue that maintaining an interpretable and persistent semantic scene structure is equally critical: not only for modelling the inspection scene but also for achieving long-term operational autonomy, as motivated in~\cite{Hughes2022HydraAR}.

To address the current challenges, the proposed framework xFLIE couples 3DLSG construction with the online inspection planner FLIE. The 3DLSG is incrementally updated based on the robot’s observations, while concurrently serving as a foundation for on-demand decision-making and path planning. Through this integration, the planner can (i) build and maintain a persistent, interpretable inspection scene representation, (ii) identify and prioritize candidate inspection targets, (iii) efficiently compute traversable routes over the abstraction layers, and (iv) interpret high-level semantic navigation commands issued by a human operator over the hierarchical scene representation. This formulation contributes towards a scalable, queryable, and robot-interpretable scene representation for large-scale autonomous missions in unknown environments.

\subsection{Contributions}~\label{sec:contributions}
Based on the above discussion, we present the main contributions of this work below.

Firstly, we introduce \textbf{3D Layered Semantic Graphs (3DLSG)}, a novel solution to construct and maintain actionable hierarchical inspection scene graphs. Prior work on 3D scene graphs defines task-specific taxonomies (e.g., $\textit{Buildings} \rightarrow \textit{Rooms} \rightarrow \textit{Places}$ for indoor SLAM) where the hierarchy is drawn from the geometry of the scene~\cite{rosinol2021kimera,bavle2022situational}. In contrast, our taxonomy reflects the planning-oriented hierarchy required for inspection missions: $\textit{Target} \rightarrow \textit{Level} \rightarrow \textit{Pose} \rightarrow \textit{Feature}$. For each inspection \textit{Target} discovered during exploration, we encode planning-relevant abstractions such as the \textit{Level} of inspection carried out, associated \textit{Pose} configurations visited, and observed \textit{Features} detected during inspection. We present detailed information on the overall structure of the 3DLSG and the internal composition of each abstraction layer in Section.~\ref{sec:3D_LSG}.

Secondly, we demonstrate the real-time construction and use of the 3DLSG during semantic inspection mission through integration with an existing planner (FLIE: First-Look Inspect-Explore~\cite{viswanathan2023towards}), resulting in a new framework: \textbf{xFLIE (Enhanced First-Look Inspect-Explore)}. Particularly, this architecture highlights the on-demand utilization of the evolving 3DLSG model by the planner for target selection, path planning and semantic navigation. While the 3DLSG is constructed by the planner, it also assists the planner by enabling informed target selection at the \textit{Target} layer and supporting hierarchical path planning across the graph’s traversable abstraction layers. Additionally, we showcase the advantages of maintaining a queryable representation by allowing human operators to issue navigation queries to the robot to highlight semantic path planning. We discuss the xFLIE framework in Section~\ref{sec:xFLIE}.

In summary, our main contributions are as follows:
\begin{enumerate}
    \item We introduce \textbf{3D Layered Semantic Graphs (3DLSG)}, a novel framework for the incremental, real-time construction of an actionable hierarchical inspection scene graph.
    
    \item We present \textbf{xFLIE}, an integrated system combining the scene graph constructor with an inspection planner, enabling semantic-aware inspection missions.
    
    \item We analyze the scalability of the 3DLSG and its impact on planning performance in large-scale environments, demonstrating efficient planning over conventional volumetric map-based approaches.
    
    \item We validate our approach through experimental evaluation in real-world outdoor urban and subterranean environments, as well as through rigorous simulations.
\end{enumerate}

An audio-visual media presenting the proposed architecture can be accessed here: \url{https://youtu.be/T-C-ZAGCH_o}.

\section{Problem Formulation}\label{sec:problem}
Let $\bm{x}_k= (\bm{x}^p_k,\bm{x}^q_k)$ denote the robot state, such that $\bm{x}^p_k \in \mathbb{R}^3$ is the position and $\bm{x}^q_k \in \mathbb{SO}(3)$ is the orientation of the robot. Let $z_k \in \mathcal{Z}$ denote the observation at time step $k$, where $\mathcal{Z}$ is the observation space. The observation $z_k$ is represented as,
\[
z_k = (\bm{z}^p_k, \bm{z}^q_k, z^l_k, z^s_k, z^a_k, z^i_k),
\]
where,
$\bm{z}^p_k \in \mathbb{R}^3$ denotes the position, 
$\bm{z}^q_k \in SO(3)$ the orientation, 
$z^l_k $ the semantic label, 
$z^s_k \in \mathbb{R}_+$ the segmentation score, 
$z^a_k \in \mathbb{R}_+$ the segmented mask area, 
and $z^i_k$ the observed RGB image.

The robot acts according to a policy $\pi \in \{\pi^{\text{expl}}, \pi^{\text{insp}}\}$ consists of two modes: 
\emph{exploration} $\pi^{\text{expl}}$ and \emph{inspection} $\pi^{\text{insp}}$ and incrementally builds a hierarchical scene graph,
\[
\mathcal{G}_k = (\mathcal{V}_k, \mathcal{E}_k, \mathcal{A}_k),
\]
representing the inspection scene observed up to time $k$. 
The graph captures both metric and semantic structure and is organized into abstraction layers,
\[
\mathbb{L} = \{\textit{Target},\, \textit{Level},\, \textit{Pose},\, \textit{Feature}\}.
\]
Each layer $l \in \mathbb{L}$ defines a local graph 
$\mathcal{G}^l_k = (\mathcal{V}^l_k, \mathcal{E}^l_k, \mathcal{A}^l_k)$,
where $\mathcal{V}^l_k$, $\mathcal{E}^l_k$, and $\mathcal{A}^l_k$ denote the nodes, edges, and node attributes at that layer, respectively. For example, the \textit{Target} layer is represented through the graph $\mathcal{G}^T_k= (\mathcal{V}^T_k,\mathcal{E}^T_k,\mathcal{A}^T_k)$, with the superscript $T$ denoting the layer association.

The nodes $\mathcal{V}^l_k$ represent semantic concepts within layer $l$, the edges $\mathcal{E}^l_k$ model spatial and symbolic topological relationships, while the attributes $\mathcal{A}^l_k$ capture metric, semantic and sensor information required for decision-making and path planning. In initially unknown environments $\mathbb{E}$, the robot operates under partial observability.  
The executed policy $\pi \in \{\pi^{\text{expl}}, \pi^{\text{insp}}\}$ consists of two modes: 
\emph{exploration} $\pi^{\text{expl}}$ and \emph{inspection} $\pi^{\text{insp}}$.
During execution of policy $\pi$, the robot observes visible semantic objects, $s_e,s_i \in S$ during exploration and inspection, respectively, from state $\bm{x}_k$ and registers them in the corresponding scene graph layer $\mathcal{G}^l_k$.

The unified hierarchical graph is then given by,
\[
\mathcal{G}_k = \bigcup_{l \in \mathbb{L}} \mathcal{G}^l_k.
\]

\textit{Problem 1: \textbf{Hierarchical Scene Graph Construction}}

Construct and maintain an unified hierarchical graph of the inspection scene that captures observed semantic objects, their relations, and spatial organization, providing sufficient actionable context for subsequent planning.

Formally, given a sequence of observations $\{z_1, \dots, z_k\}$ obtained under policy $\pi$, the layers of the hierarchical scene graph is updated incrementally as

\begin{equation}\label{eq:graph_update_operator}
\begin{aligned}
    \mathcal{G}^l_k &= \mathcal{R}(\mathcal{G}^l_{k-1}, z_k)\, \\
    \mathcal{G}_k &= (\mathcal{V}_{k-1} \cup \mathcal{V}^l_{k}, \mathcal{E}_{k-1} \cup \mathcal{E}^l_{k},\mathcal{A}_{k-1} \cup \mathcal{A}^l_{k})
\end{aligned}
\end{equation}
where $\mathcal{R}(\cdot)$ denotes the operator that registers new observation $z_k$ into the existing layer graph $\mathcal{G}^l_{k-1}$. The operator $\cup$ defines the incremental merging of the vertex, edge and attribute sets.

\begin{figure*}[b]
    \centering
    \includegraphics[width=\linewidth]{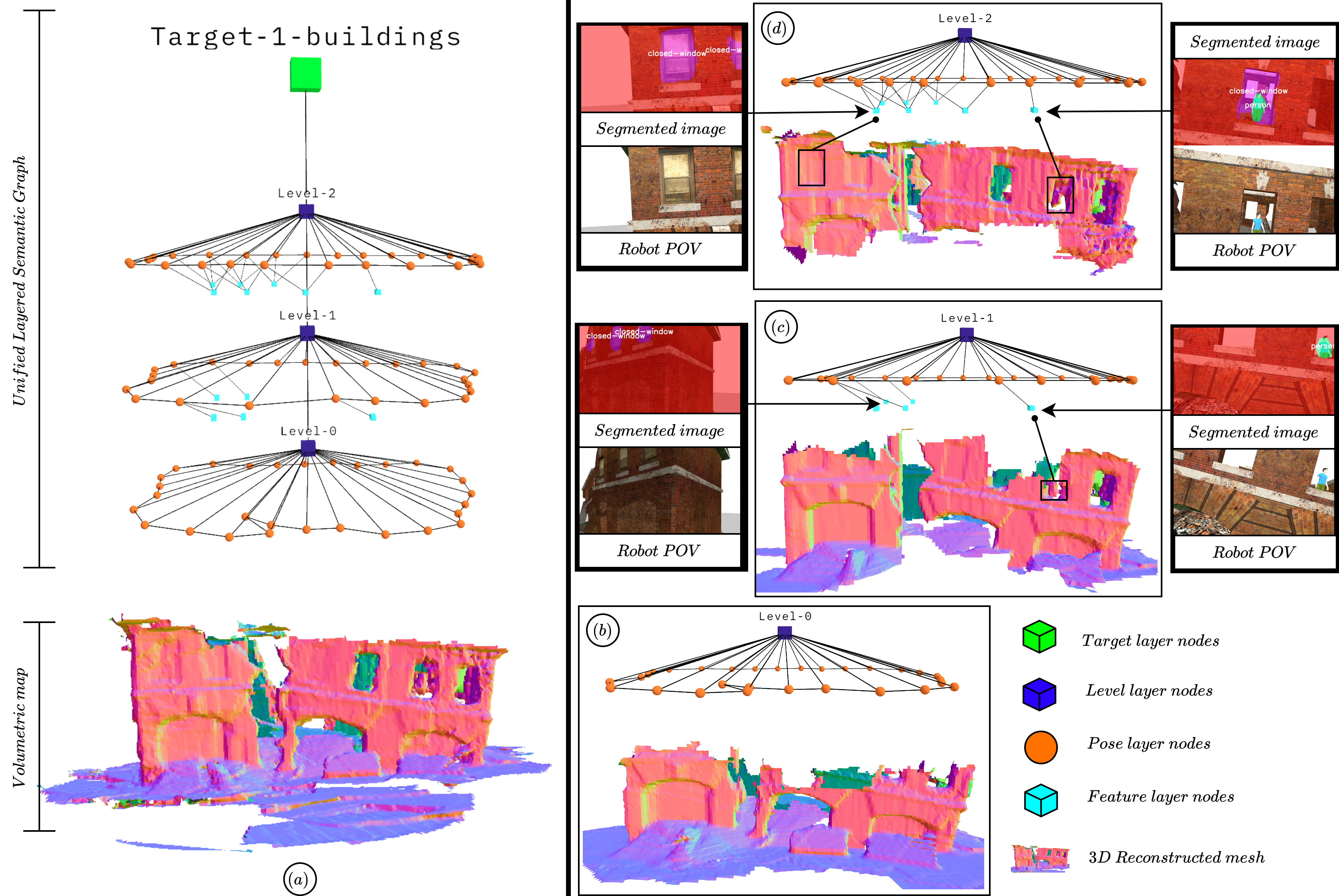}
    \caption{
    Example of a 3D Layered Scene Graph (3DLSG) constructed by the \textit{xFLIE} architecture during a building inspection scenario.
    The system segments and encodes semantic features such as \textit{person} and \textit{closed-window} within the volumetric map~\cite{oleynikova2017voxblox}, resulting in a unified multi-layer representation of the scene.}
    \label{fig:xflie_r2_single_building_inspection}
\end{figure*}
\textit{Problem 2: \textbf{Semantic-Aware Inspection}}

Utilize the scene graph $\mathcal{G}_k$ to identify high-utility inspection targets and plan feasible paths towards it to initiate inspection. At the conclusion of exploration policy $\pi^{\text{expl}}$, the planner selects the most promising target node $v_{D*} \subseteq \mathcal{V}^T_k$ based on a utility measure $\mathcal{U}(\cdot)$ and plans a traversable route toward it using a hierarchical path-planner $\Pi(\cdot)$ over the current 3DLSG model $\mathcal{G}_k$.

\begin{align}
    &\pi^{\text{insp}}(v_{D*}) \\
    \text{such that,} \quad
    &\|\bm{x}^p_k - \bm{p}(v_{D*})\| \leq \epsilon, \nonumber\\
    &\Pi(v_{D*}, \mathcal{G}_k),\nonumber\\
    &v_{D*} = \underset{v_D \in \mathcal{V}^T_k}{\arg\max} \; \mathcal{U}(v_D), \nonumber
\end{align}

where $\mathcal{U}(\cdot)$ evaluates the inspection utility of each candidate target,
and $\Pi(\cdot)$ denotes the hierarchical path-planning operator that computes a feasible route over $\mathcal{G}_k$ to reach $v_{D*}$ and $\bm{p}(v)$ maps a vertex $v \in \mathcal{V}^l_k$ to its positional coordinates.
The inspection policy $\pi^{\text{insp}}$ is initiated when the robot's position $\bm{x}^p_k$ satisfies the proximity condition 
$\|\bm{x}^p_k - \bm{p}(v_{D*})\| \leq \epsilon$, where $\epsilon \in \mathbb{R}_+ $ is the minimum distance to the target node.

\textit{Terminologies:} We distinguish between the terms, \textit{semantic targets} and \textit{semantic features}, as follows: (a) \textit{semantic targets} refer to targets of interest, such as \textit{car}, \textit{house}, or \textit{truck}, that is observed in the environment as candidates for further inspection; and (b) \textit{semantic features} refer to the observed and segmented features of a semantic target, such as the \textit{window} or \textit{door} of a house, or the \textit{hood} of a car.

\section{3D Layered Semantic Graph}\label{sec:3D_LSG}
A 3D Layered Scene Graph (3DLSG) is a hierarchical graph designed to incrementally model the metric, semantic, and relational structure of the scene as the robot inspects, while concurrently supporting path planning and decision-making over the evolving representation. The 3DLSG is organized into four layers: \textit{Target}, \textit{Level}, \textit{Pose}, and \textit{Feature}, grounded within the domain of an inspection mission. It is presented in the decreasing order of abstraction: from a global perspective representing the inspected targets and the connectivity between them (\textit{Target}) to local inspection behaviour representing the altitude changes (\textit{Level}), the tracked viewpoints at each level (\textit{Pose}) and observed semantic features at each viewpoint (\textit{Feature}).

Each layer maintains a graph that represents semantic concepts as nodes and intra-layer edges capturing their relational structure. Additionally, it encodes node attributes derived from observations during the mission. The edges capture spatial and symbolic relations between nodes. The spatial edges encode Euclidean distance between coordinates as attributes to facilitate traversable route planning, and the symbolic edges model the associative connectivity of the hierarchy to support scene awareness.

Figure~\ref{fig:xflie_r2_single_building_inspection} illustrates 3DLSG constructed for inspecting an \textit{a priori} unknown multi-storey building in a simulated environment. In this scenario, the robot first explored the environment and registered the detected semantic targets in the \textit{Target} layer. Once a specific target object was selected for inspection, the robot progressively populates the subsequent layers. During the inspection phase (refer Fig~\ref{fig:xflie_r2_single_building_inspection}(b)–(d)), the \textit{Level} layer is populated with the levels of inspection corresponding to planned altitude changes from the planner. The \textit{Pose} layer encodes the tracked viewpoints associated with each level, while the \textit{Feature} layer registers the detected semantic features (e.g., windows, doors, or façade details) observed from each viewpoint.

\subsection{Layer Definition and Composition}~\label{sec:layer_construction}
\begin{figure}
    \centering
    \includegraphics[width=\linewidth]{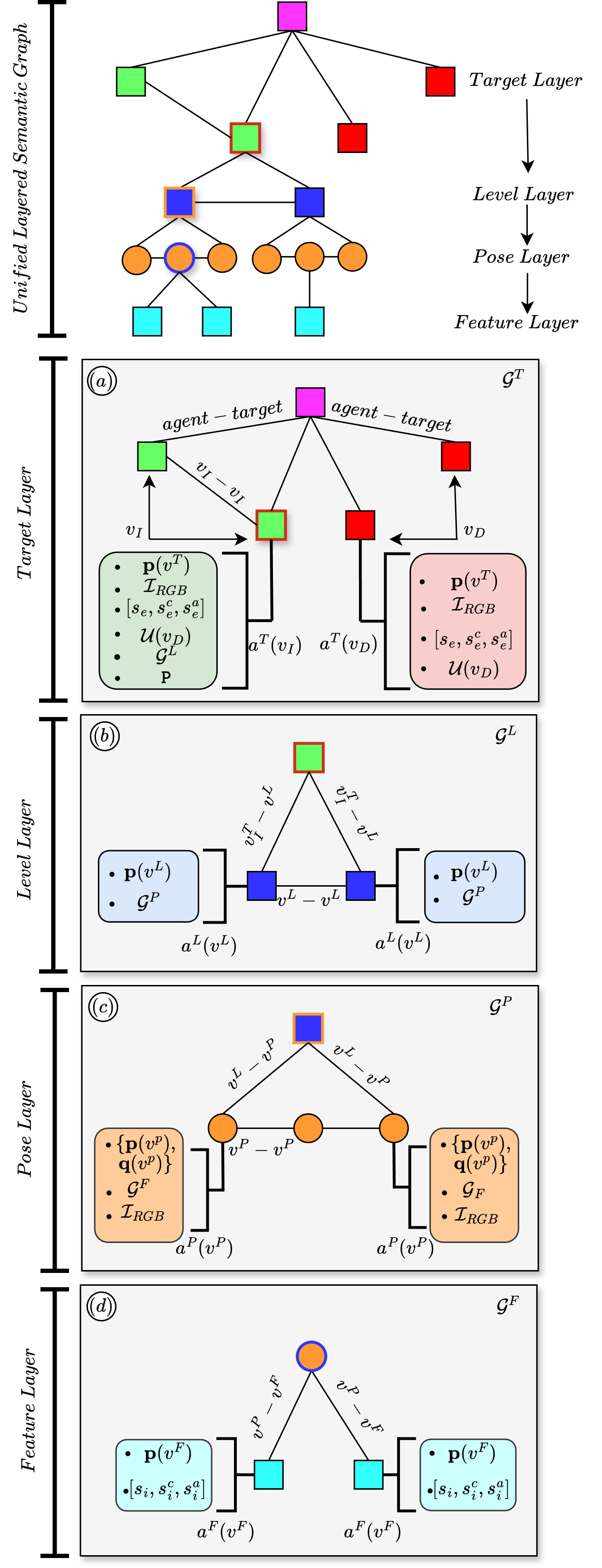}
    \caption{A schematic representation of the internal composition of the 3DLSG.}
    \label{fig:LSG_explantory_figures}
\end{figure}
\textbf{Target Layer}: Semantic targets for inspection $s_e \in S$ are registered as nodes $v^{T} \in \mathcal{V}^T_k$ within the \textit{Target}-layer graph $\mathcal{G}_k^T$. We model the parent node within the $Target$ layer as a representation of the robotic agent encoded with information on the current robot state $\bm{x}_{k}=(\bm{x}^p_k,\bm{x}^q_k)$. The child nodes within this graph can be differentiated into two main subsets, $\mathcal{V}^T_{k,I},\mathcal{V}^T_{k,D} \subseteq \mathcal{V}^T_k$, where $v_I \in \mathcal{V}^T_{k,I}$ denotes the set of target nodes that have been inspected and $v_D \in \mathcal{V}_{k,D}^T$ denoting the set of nodes that have been detected and are remaining to be inspected.
Figure.~\ref{fig:LSG_explantory_figures}(a) presents the internal composition of \textit{Target} layer graph $\mathcal{G}^T_k$.

Intra-layer edges $\mathcal{E}^T_k$ model connections symbolizing (i) parent-child relations between the registered inspection targets and the robot, and (ii) spatial relations between the inspected target nodes $v_{I}$-$v_{I}$. Each node encodes the attribute $a^T_k \in A^T_k$ from the observation $z_k$, consisting of (i) the estimated 3D location of the target node defined from $\bm{z}^p_k$, (ii) RGB image information, $\mathcal{I}_{RGB}$, of the target from $z^i_k$, (iii) semantic label $s_e$ from $z^l_k$, (iv) segmentation confidence score $s^c_e$ from $z^s_k$, (v) area of segmentation mask $s^a_e$ from $z_k^{a}$, and (vi) a utility heuristic $U \in \mathbb{R}_+$ of the semantic for inspection target selection.

Upon inspection of a detected target node $v_D$, the corresponding attribute $a^T_k$ is augmented with the \textit{Level} layer graph $\mathcal{G}^L_k$ and its corresponding \textit{Pose} and \textit{Feature} layer graphs, $\mathcal{G}^P_k$ and $\mathcal{G}^F_k$, respectively. Additionally, a simple polygon $\texttt{P} \in \mathbb{R}^2$, representing the occupied region of the target is appended to the node attribute at the end of inspection. For detailed information on the $\bm{\texttt{P}}$ construction and its use, the readers are directed to Sec.~\ref{sec:xFLIE}-\ref{sec:xflie_insp}.

\textbf{Level Layer}: The \textit{Level}-layer graph, $\mathcal{G}^L_k$, associated with each target node in the \textit{Target} layer, consists of child nodes $v^L \in \mathcal{V}^L_k$. These nodes represent the distinct inspection levels executed for the current target (e.g., \textit{Level}-0, \textit{Level}-1). A new node is instantiated either at the start of a target's inspection or when the planner commands a transition to the next inspection altitude. For details on establishing inspection levels, see~\cite{viswanathan2023towards}.

Each level node $v^L$ is encoded with attributes $a^L_k \in \mathcal{A}^L_k$ derived from the planner-based observation $z_k$. These include (i) the 3D position of the level node defined from $\bm{z}^p_k$, and (ii) the associated \textit{Pose}-layer graph $\mathcal{G}^P_k$, which represents the tracked view-poses under the current inspection level node $v^L_{curr}$. The current target node $v^T_{curr}$ being inspected serves as the parent node within this subgraph. Symbolic intra-layer connections link the parent node to its child level nodes, while edges between adjacent $v^L$ nodes are assigned weights equal to the Euclidean distance between their 3D coordinates.

Fig.~\ref{fig:LSG_explantory_figures}(b) illustrates the composition of the \textit{Level}-layer graph and its registered attributes, populated during the inspection of the active target node (highlighted in red).

\begin{figure*}[hb]
    \centering
    \includegraphics[width=\linewidth]{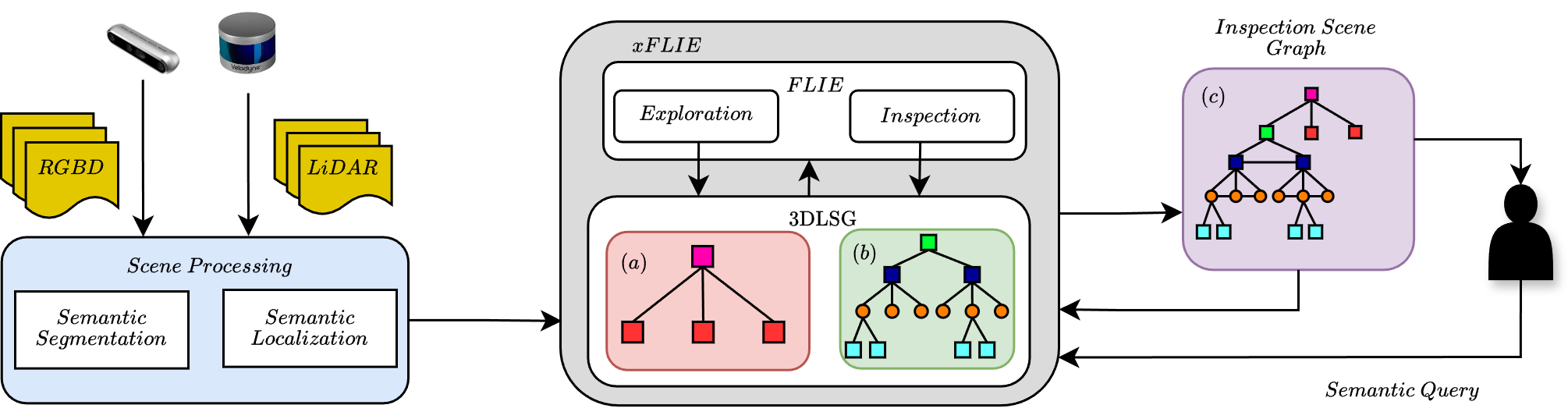}
    \caption{The proposed architecture xFLIE which integrates FLIE, an inspection and exploration planner, with the 3DLSG, an actionable hierarchical scene representation, for semantic-aware inspection in unknown environments. xFLIE uses RGB, Depth and LiDAR measurements (shown on \textit{top-left}) to segment and localize desired semantics during inspection and exploration. The bifurcated FLIE planner populates the corresponding layers, (a) $\textit{Target}$ layer during exploration and (b) $\textit{Level},\textit{Pose},\textit{Feature}$ layers during inspection, within the 3DLSG and outputs (c) an actionable hierarchical scene representation which is used for addressing planning and semantic queries.}
    \label{fig:xflie_framework}
\end{figure*}

\textbf{Pose Layer}: The \textit{Pose}-layer graph, $\mathcal{G}^P_k$, defined under the current inspection level $v^L_{curr}$, consists of child nodes $v^P \in \mathcal{V}^P_k$ representing the view-poses tracked during inspection. Each node is encoded with attributes $a^P_k \in \mathcal{A}^P_k$ derived from the planner-provided observation $z_k$, including (i) the tracked view-pose state $(\bm{p}(v^P),\bm{q}(v^P))$ defined from $\bm{z}^p_k$ and $\bm{z}^q_k)$, where $\bm{q}(v)$ maps a vertex to its orientation (ii) the RGB image information captured at the commanded view-pose $z^i_k$, and (iii) the associated \textit{Feature}-layer graph $\mathcal{G}^F_k$, which represents the observed semantic features from the current viewpoint.

The current inspection level node serves as the parent node of this subgraph and is connected to the initial and final pose nodes registered during the inspection of the current level (highlighted in \textit{orange}). Edges within $\mathcal{G}^P_k$ are weighted by the Euclidean distance between the corresponding 3D spatial coordinates, and each pose node is additionally connected to its adjacent pose node through weighted edges.

Fig.~\ref{fig:LSG_explantory_figures}(c) illustrates the internal composition of $\mathcal{G}^P_k$, including the registered attributes and connectivity established under an inspection level node.

\textbf{Feature Layer}: The \textit{Feature}-layer graph, $\mathcal{G}^F_k$, registers the observed semantic features $s_i \in \mathcal{S}$ of the target node during inspection. Each feature node is encoded with attributes $a^F_k \in \mathcal{A}^F_k$ derived from the observation $z_k$, including (i) the 3D position of the feature defined from $\bm{z}^p_k$, (ii) its semantic label $s_i$ from $z^l_k$, (iii) the segmentation confidence score $s^c_i$ from $z^s_k$, and (iv) the segmented mask area $s^a_i$ from $z^a_k$. Each feature node shares a symbolic edge with the parent pose node from which it was observed, representing the relational link between a viewpoint and its detected semantic entities.

Fig.~\ref{fig:LSG_explantory_figures}(d) illustrates $\mathcal{G}^F_k$, populated with the observed semantic features and their registered attributes under the current inspection view-pose node (highlighted in \textit{red}).

\section{xFLIE: Leveraging Hierarchical Representation for Autonomous Inspection}\label{sec:xFLIE}

In this section, we discuss in detail the implementation and deployment of xFLIE architecture towards addressing semantic-aware inspection missions in unknown environments. The xFLIE architecture comprises of a modified version of the inspection framework FLIE~\cite{viswanathan2023towards} and the 3DLSG constructor, both working in tandem to manage mission objectives. Figure.~\ref{fig:xflie_framework} provides an insight of the implemented pipeline, integrating FLIE and 3DLSG, to address semantic-aware inspection in unknown environments with the construction of a hierarchical inspection scene graph.

In general, FLIE framework exhibits an \textit{explore-inspect-explore} approach until there remains no further observable semantic targets within the operating environment. In this work, we modify the original \textit{exploration} planner to perform two exploration policies : (a) the initial $360^\circ$ survey of the immediate vicinity of the robot denoted as $\pi^{expl:360}$ and (b) the local exploration of the region around the inspected object denoted as $\pi^{expl:LE}$. We denote the inspection policy as $\pi^{insp}$ that would be executed for each registered semantic target. The readers are directed to the work in~\cite{viswanathan2023towards} for detailed information on the planning aspect of FLIE.

\subsection{Exploration}\label{sec:xflie_expl}

\textbf{Target Layer Population:} During exploration policy $\pi^{expl}$, $\mathcal{G}^T_{k-1}$ is populated with the current observation $z_k$ from the environment to obtain an unoptimized representation $ \Tilde{\mathcal{G}}^T_k$ as in~\eqref{eqn:graph_update}.
\begin{equation}\label{eqn:graph_update}
    \Tilde{\mathcal{G}}^T_k=\mathcal{R}(\mathcal{G}^T_{k-1}, z_k).
\end{equation}
where, $\mathcal{R}(\cdot)$ denotes the node registration operation. For each detected target node $v_{D,i}$ registered during exploration, we optimize $\mathcal{\Tilde{G}}_T$ as in~\eqref{eqn:graph_optim}.
\begin{equation}
\begin{aligned}\label{eqn:graph_optim}
    &\mathcal{G}^T_k = \Phi( \Tilde{\mathcal{G}}^T_k,v_{D,i}),\\
    \text{s.t.}\\
& \mathcal{D}\!\bigl(\bm{p}(v_{D,i}),\, \bm{p}(v_{D,j})\bigr) > d_{\mathrm{check}}^{T},\forall\,v_{D,i}\neq v_{D,j}\in \mathcal{V}_D^{T},\\
& \mathcal{P}\!\left(v_{D,i},\, v_{I,m}\right) = 0,
\quad \forall\, v_{I,m} \in \mathcal{V}_{I,\mathrm{wsr}}^{T}, \\
& \mathcal{V}_{I,\mathrm{wsr}}^{T} 
= \bigl\{ v_I \in \Tilde{\mathcal{V}}^T_{I} : \mathcal{D}(\bm{p}(v_{D,i}),\bm{x}^p_k)\le d_{\max},\ \theta_{v_I}\le \alpha \bigr\}.
\end{aligned}
\end{equation}

where, $\Phi(\cdot)$ denote the optimization operation on $\mathcal{G}^T_{k-1}$, $\mathcal{P}(\cdot)$ is the point-in-polygon evaluation defined as,
\begin{align}\nonumber
    \mathcal{P}\!\left(v_{D,i},\, v_{I,m}\right) = \begin{cases}
        1, \text{if}\ \bm{p}(v_{D,i}) \in \texttt{P}_{v_{I,m}},\\
        0, \text{otherwise}
   \end{cases}
\end{align}
and $\mathcal{V}^{I,\mathrm{wsr}}_{k}$ denotes the subsampled set of inspected target nodes within sensing range from the current position of the robot $\bm{x}^p_k$, $d_{max} \in \mathbb{R}_+$ is the maximum reliable sensor range, $\alpha \in \mathbb{R}_+$ is the Field-of-View (FOV) of the robot, $\mathcal{D}(\cdot)$ denotes the Euclidean distance evaluation of the input positional coordinates, $\theta_{v_I}$ is the viewing angle under which a target node is seen by the robot and $d^T_{check} \in \mathbb{R}_+$ is the minimum validation distance for a node to be considered unique.

\begin{figure}[htpb]
    \centering
    \includegraphics[width=0.8\linewidth]{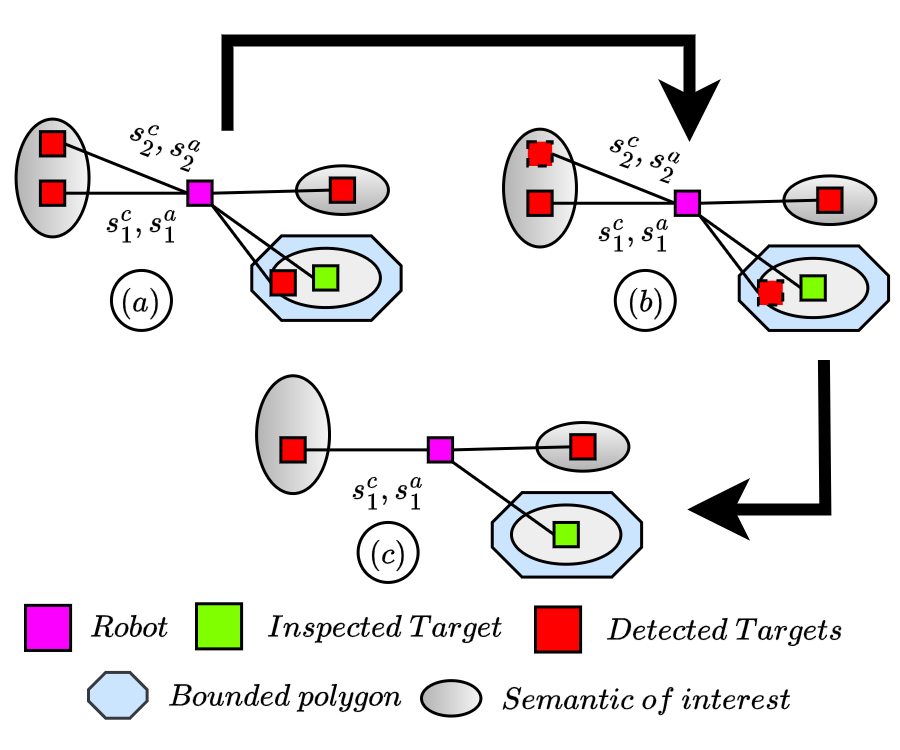}
    \caption{An overview of the optimization process of \textit{Target} layer graph after the exploration phase.}
    \label{fig:TargetGraphOptimization}
\end{figure}

Figure.~\ref{fig:TargetGraphOptimization} presents an overview of the modelled optimization of the initial $\mathcal{G}^T_k$ during the exploration phase of the planner. In Fig~\ref{fig:TargetGraphOptimization}(a), the unoptimized $\mathcal{\Tilde{G}}^T_k$ graph comprising of detected and inspected nodes is presented. The invalid/duplicate nodes are highlighted with hashed-boundary lines in Fig.~\ref{fig:TargetGraphOptimization}(b). Figure.~\ref{fig:TargetGraphOptimization}(c) presents the optimized \textit{Target}-layer graph $\mathcal{G}^T_k$ for subsequent evaluation of the decision to inspect. 

At the end of $\pi^{\text{expl}}$, the optimized \textit{Target}-layer graph $\mathcal{G}^T_k$ is merged with the unified 3DLSG model $\mathcal{G}_{k-1}$ to update the overall scene graph, yielding $\mathcal{G}_k$ as defined in~\eqref{eqn:target_layer_merging}.
\begin{equation}\label{eqn:target_layer_merging}
    \mathcal{G}_k = (\mathcal{V}_{k-1} \cup \mathcal{V}^T_{k}, \mathcal{E}_{k-1} \cup \mathcal{E}^T_{k},\mathcal{A}_{k-1} \cup \mathcal{A}^T_{k} )
\end{equation}
where, $\cup$ denotes the graph union operation that merges the vertex, edge, and attribute sets.

Subsequent to~\eqref{eqn:target_layer_merging}, we evaluate the utility $\mathcal{U}$ of the detected target nodes as a weighted sum of three parameters: (a) the relevance of node, which in this work is considered to be the pixel area of the segmentation with respect to image resolution $A(v_{D}) \in \mathbb{R}_{+}$. Therefore, larger semantic targets are prioritized to be inspected before smaller ones. (b) spatial proximity to other neighbouring target nodes $N(v_{D}) \in \mathbb{R}_{+}$, positively biasing nodes which are more central to other target nodes and (c) spatial proximity of the node to the robot $P(v_{D}) \in \mathbb{R}_{+}$.

Equation~\eqref{eqn:utility_heuristic} presents the mathematical formulation of the utility heuristic function implemented to rank semantic targets for further inspection. At the instant of determining the next inspection target,~\eqref{eqn:utility_heuristic} is evaluated at runtime and the candidate node $v_{D*}$ having the maximum utility is then selected for inspection.
\begin{align}\label{eqn:utility_heuristic}
    v_{D*} &= \underset{v_D \in \mathcal{V}_D^T}{\operatorname{argmax}} \quad \mathcal{U}(v_D) \\
    \mathcal{U}(v_{D}) &=   S_pP(v_{D}) + S_aA(v_{D}) + S_nN(v_{D}) \nonumber
\end{align}
such that,
\begin{gather*}\nonumber
    P(v_D) = \frac{1}{||\bm{x}^p_{k} - \bm{p}(v_{D})}||\\ \nonumber
    A(v_D) = \frac{s^{a}_{v_D}}{I_{w}*I_{h}}\\ \nonumber
    N(v_D) = \frac{1}{\Bar{d}_{v_D}}\nonumber\\
    \Bar{d}_{v_D} = \frac{1}{|\mathcal{V}_D^T|-1} \sum_{j=1,j\neq i}^{|\mathcal{V}_D^T\}|-1} ||\bm{p}(v_{D,i}) - \bm{p}(v_{D,j})||\nonumber
\end{gather*}
where, $s^{a}_{v_D}$ is the area of the segmentation mask, $I_w,I_h$ being the width and height of the RGB image and $\Bar{d}_{v_D}$ is average distance of the current evaluated node $v_{D,i}$ to other uninspected target nodes $v_{D}$. $S_p,S_a,S_n \in \mathbb{R}_+$ are the weights corresponding to the three modelled parameters which influence the inspection decision.

Finally, a traversable route is planned on the current 3DLSG representation $\mathcal{G}_k$ to reach the target for inspection. We provide the details on the implemented hierarchical path planner in Section.~\ref{sec:xFLIE}-\ref{sec:hpp_spp}.

\subsection{Inspection}\label{sec:xflie_insp}
During inspection of $v_{D*}$, the lower three layers, i.e $\textit{Level},~\textit{Pose}$ and $\textit{Feature}$ are initialized and populated. We discuss the method of populating each layer below.

\textbf{Level Layer Population:} The \textit{Level}-layer graph $\mathcal{G}^L_k$ is populated according to the commanded altitude changes issued by the planner during the inspection of a semantic target. Similar to~\eqref{eqn:graph_update}, the \textit{Level}-layer graph is updated with new observation and merged with unified scene graph as defined below in~\eqref{eqn:level_layer_graph}.
\begin{equation}
\begin{aligned}\label{eqn:level_layer_graph}
    \mathcal{G}^L_k &= \mathcal{R}(\mathcal{G}^L_{k-1},z_k)\\
    \mathcal{G}_k &= (\mathcal{V}_{k-1} \cup \mathcal{V}^L_{k}, \mathcal{E}_{k-1} \cup \mathcal{E}^L_{k},\mathcal{A}_{k-1} \cup \mathcal{A}^L_{k})
\end{aligned}
\end{equation}
where the observation $z_k$, as defined earlier for the \textit{Level}-layer in Section~\ref{sec:3D_LSG}, is derived from the planner behaviour.

Since FLIE is a map-free, reactive view-planner, it leverages RGB images from tracked view-poses to recognize previously inspected surfaces. When the robot revisits a previously inspected surface within the current level of inspection, the planner advances to inspect the next level. We denote $\bm{\mathcal{I}}^c = \{i^c_1,i^c_2,...,i^c_N\}$ as the set of candidate images captured for an observation horizon ($N \in \mathbb{Z}_+$). Additionally, we define $d^I_{thresh} \in \mathbb{R}_+$ as the minimum distance between the point of initialization of the level of inspection and the point from which the scene similarity score, ($\Gamma \in \mathbb{R}_+$), is computed. The spatial-proximity check is done to limit the use of GPU-resources by the feature matcher during evaluation of scene similarity, which alternatively would run after the initial database of images collected.

For each candidate image $i^c \in \bm{\mathcal{I}}^c$, we denote the matched local features between the query image $\mathcal{I}_q$ and the candidate image $i^c$ by $\mathcal{M}^{i^c}_{\mathcal{I}_q}$.
\begin{figure}[htbp]
    \centering
    \includegraphics[width=0.8\linewidth]{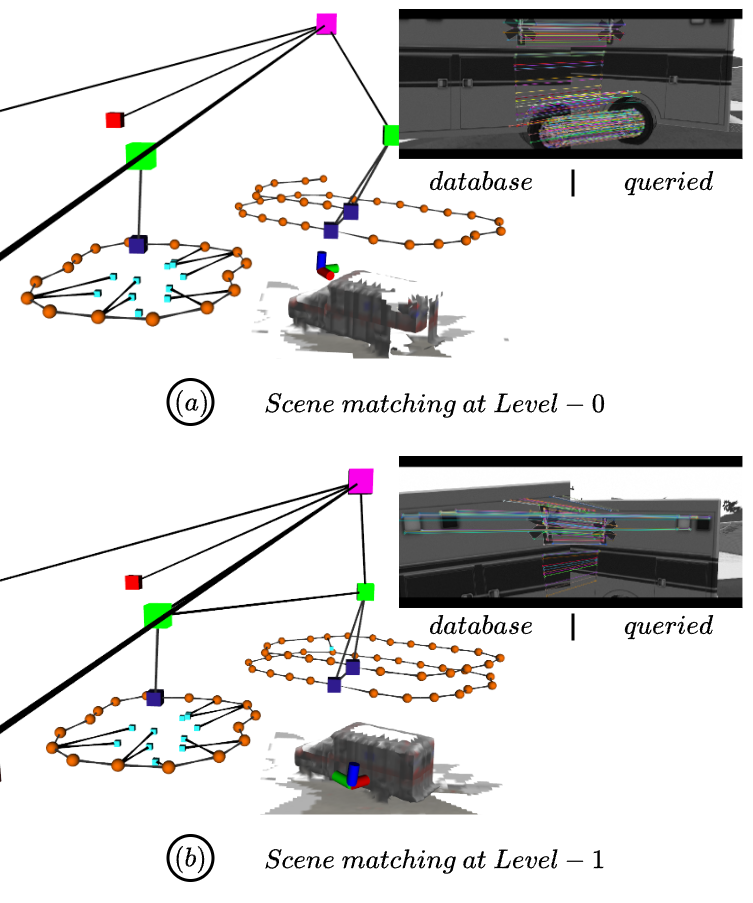}
    \caption{A runtime insight into the performance of the scene recognition module executed during inspection in a simulated urban environment.}
    \label{fig:scene_matching}
\end{figure}
We denote the set of extracted keypoints for a query image as $\mathcal{K}_q \in \mathbb{R}^2$. Given a set of candidate images $\bm{\mathcal{I}}^c = \{ i^c_n \}_{n=1}^{N}$, we compute the total number of relative matches aggregated over the horizon relative to to the total number of query keypoints. Equation~\eqref{eqn:scene_sim} formally defines the implemented scene-similarity scoring between the database of candidate images and the query image.
\begin{align}\label{eqn:scene_sim}
    \Gamma = 
\begin{cases}
    \frac{\sum^{N}_{n=1} \mathcal{M}^{i^c_n}_{\mathcal{I}_q}}{|\mathcal{K}_q|},& if d^I_{curr}\leq d^I_{thresh}\\
    0,              & otherwise
\end{cases}
\end{align}
such that, 
\begin{equation*}\nonumber
    d^I_{curr}=||\bm{x}^p_{k}-\bm{p}(v^{L}_{curr})||
\end{equation*}
Figure.~\ref{fig:scene_matching} presents a runtime snapshot of the performance of scene matching during a simulated inspection mission. Figure.~\ref{fig:scene_matching}(a) demonstrates the matched features (colored lines) across the current and the image database populated during inspection at $\textit{Level}-0$ of a $``\textit{truck}"$ semantic. After successful recognition, the inspection planner proceeds to inspect the next level subject to meeting the vertical photogrammetric requirements defined before the mission start. In Fig.~\ref{fig:scene_matching}(b), the inspection planner reevaluates scene recognition with the updated image database for $\textit{Level}-1$.

\textbf{Pose Layer Population:} Once the \textit{Level}-layer graph $\mathcal{G}^L_k$ is initialized and populated, the corresponding \textit{Pose}-layer graph $\mathcal{G}^P_k$ is initialized under the current level node $v^L_{curr}$. The graph $\mathcal{G}^P_k$ registers the tracked view-pose configuration as nodes with corresponding attributes, as defined in Sec.~\ref{sec:3D_LSG}-\ref{sec:layer_construction}. The Pose-layer graph is updated during inspection as shown in~\eqref{eqn:pose_layer_update},
\begin{equation}
\begin{aligned}\label{eqn:pose_layer_update}
    \mathcal{G}^P_k &= \mathcal{R}( \mathcal{G}^P_{k-1},z_k) \\
     \mathcal{G}_k &= (\mathcal{V}_{k-1} \cup \mathcal{V}^P_{k}, \mathcal{E}_{k-1} \cup \mathcal{E}^P_{k},\mathcal{A}_{k-1} \cup \mathcal{A}^P_{k})
\end{aligned}
\end{equation}
where the observation $z_k$, as defined earlier for the \textit{Pose}-layer in Section~\ref{sec:3D_LSG}, is derived from the planner behaviour.

\textbf{Feature Layer Population:} The \textit{Feature}-layer graph $\mathcal{G}^F_k$ is populated by extracting the segmented semantics observed from the current inspection view-pose $v^P_{curr}$. Semantic segmentation under real-world conditions is not ideal. Additionally, a semantic feature can be observed from multiple view-poses throughout the inspection. To account for this, we construct an initial graph representation $\Tilde{\mathcal{G}}^{PF}= (\mathcal{V}^{PF},\mathcal{E}^{PF})$. Here, node set $\mathcal{V}^{PF}$ represents pairwise pose-feature $v^P-v^F$
observation and the edges $\mathcal{E}^{PF}$ represents the connection between them.

\begin{figure}[htbp]
    \centering
    \includegraphics[width=0.8\linewidth]{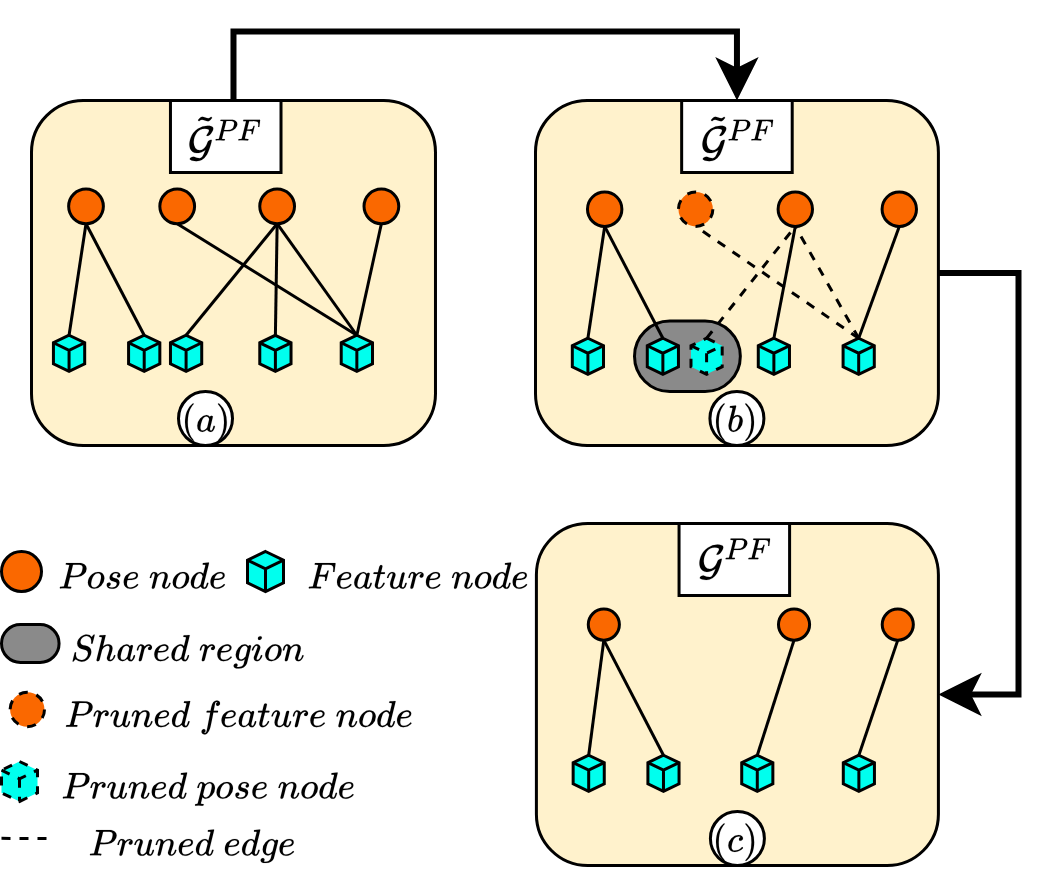}
    \caption{Visualization of the optimization of the interim $\mathcal{\Tilde{G}}^{PF}$ during inspection of a semantic target.}
    \label{fig:graph_pf_optimization}
\end{figure}

During inspection, input observations are registered and $\Tilde{\mathcal{G}}^{PF}$ is optimized at the end of each inspection level to prune low-quality or redundant observations, as  formulated in~\eqref{eqn:feature_layer_graph_optim}
\begin{equation}\label{eqn:feature_layer_graph_optim}
\begin{aligned}
    &\mathcal{G}^{PF} = \Phi(\Tilde{\mathcal{G}}^{PF}),\\
    \text{s.t.}\quad
    &\mathcal{D}\bigl(\mathbf{p}(v_i^{F}),\, \mathbf{p}(v_j^{F})\bigr) > d_{\mathrm{check}}^{F},
      \forall\, v_i^{F}, v_j^{F}\!\in\!\mathcal{V}^{F},\ i\neq j, \\
    &v^{P^\star}(v^{F}) \in 
      \operatorname*{arg\,max}_{\,v^{P}:\,(v^{P},v^{F})\in\mathcal{V}^{PF}} 
      c~\!\bigl(v^{P},v^{F}\bigr),
      \forall\, v^{F}\in\mathcal{V}^{F}.
\end{aligned}
\end{equation}
where, $c~\!\bigl(v^{P},v^{F}\bigr)$ is the associated segmentation confidence $s^c_i$ of the semantic feature $v^F$ observed from the view-pose $v^P$ and $d^F_{check} \in \mathbb{R}_+$ is the minimum required distance between any two semantic features. Figure.~\ref{fig:graph_pf_optimization} presents a visual representation of the pruning process undertaken to optimize $\mathcal{\Tilde{G}}_{PF}$. Figure.~\ref{fig:graph_pf_optimization}(a) presents the initial $\mathcal{\Tilde{G}}^{PF}$ graph populated with semantic features with the corresponding view-pose during inspection.

The optimized $\mathcal{G}^{PF}$ is then merged into the unified scene graph to yield an updated global inspection scene graph. Figure~\ref{fig:graph_pf_optimization}(c) presents the optimized $\mathcal{G}^{PF}$ graph after the pruning process.

\textbf{Containment Polygon Construction:} After inspecting a target, a polygon $\bm{\texttt{P}}^{v_{I}^T}$ is computed using the registered pose nodes $\mathcal{V}^P$ of the first inspection level node. Each polygon vertex corresponds to a pose node whose position matches the node's positional attributes. This assists the framework in validating targets detected during exploration. In our previous work, we implemented an augmented descriptor database in in conjunction with SIFT~\cite{lowe1999object} to recognize previously inspected targets. However, this approach was inherently sensitive to view-pose variance and performance degraded under dynamic outdoor lighting conditions. A geometric check, on the other hand, offered a deterministic approach independent of environmental conditions and appearance-based evaluations.

\subsection{Hierarchical and Semantic Path Planning}\label{sec:hpp_spp}
\begin{figure*}[hbtp]
    \centering
    \includegraphics[width=0.9\linewidth]{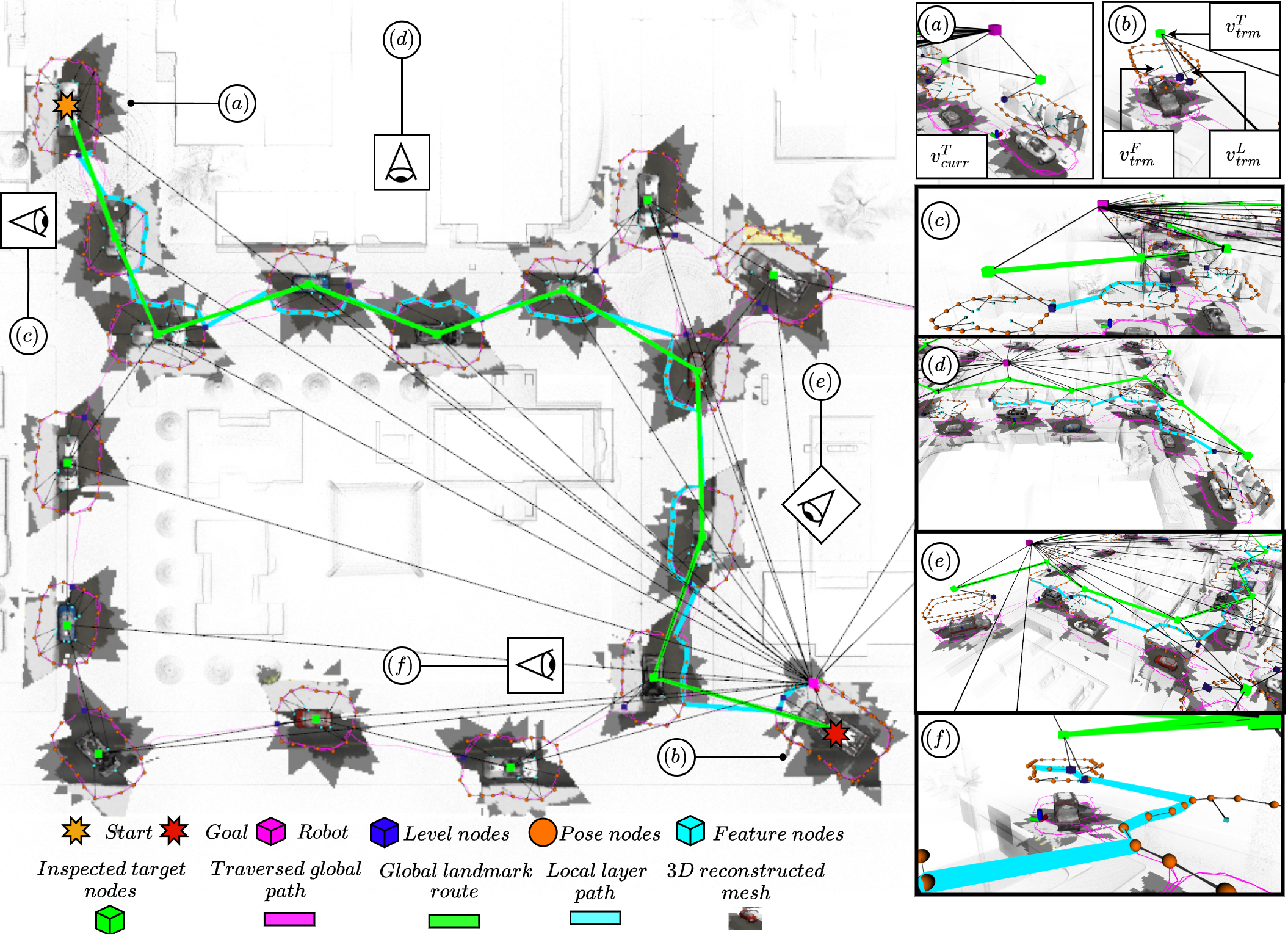}
    \caption{An instance of the response from the hierarchical path planner for a semantic query at the end of the inspection mission.}
    \label{fig:HPP_resultsemQ_upd_collage}
\end{figure*}
\textbf{Hierarchical Path Planning:} A key advantage of a hierarchical environmental representation is its ability to enable efficient planning~\cite{Hughes2022HydraAR,agia2022taskography,ray2024task} compared to conventional volumetric methods. In this work, we leverage a 3D hierarchical representation for path-planning tasks, extracting traversable paths across various layers. We also show how user-queried semantic tasks can be mapped into hierarchical planning responses, taking advantage of abstract concepts embedded within the scene representation. Instead of querying the robot to navigate to a specific metric coordinate, users can issue commands based on semantic concepts, such as instructing the robot to visit a semantic feature like \textit{hood}-1 in \textit{Level}-0 of an inspected target (e.g., \textit{car}-1). 

The hierarchical planner generates traversable paths by manipulating the 3DLSG structure in response to planning queries. The navigation queries can originate either from the integrated FLIE autonomy stack, e.g., navigating to $v_{D*}$ for inspection, or from a human operator, e.g., observing a specific target component such as \textit{hood}-1. In this work, we employ Dijkstra's algorithm~\cite{dijkstra1959note} for path planning over the 3DLSG.

When a query is received, the planner first localizes the robot within the abstraction layers, obtaining the nodes $v_{curr}^T$, $v_{curr}^L$, and $v_{curr}^P$, representing the current position of the robot within the 3DLSG. Next, a global high-level route $\Pi({v_{curr}^T,v^T_{trm}})$ is computed within the \textit{Target}-layer graph $\mathcal{G}^T_k$, with the terminal node $v_{trm}^T$ defined based on the query.

The resulting global route provides a sequence of target nodes connecting the robot's current target to the destination target node. This route cannot be tracked directly since each node in the route represents the position of a target, which is in occupied space. Therefore, the planner decomposes it into segments, each segment corresponding to a traversable local path planned within the local graphs of the current target node $v_{curr}^{T}$, i.e., over the Level and Pose-layer graphs, $\mathcal{G}^L_k$ and $\mathcal{G}^P_k$, respectively. This iterative process continues until the robot reaches the target node.

\textbf{Semantic Path Planning:} 3D LSGs maintains an intuitive scene representation of semantic concepts over multiple layers of abstraction. This makes it a powerful tool that can be leveraged to address semantic tasks which can extend the lifetime of an inspection mission. Using the registered semantic labels, an operator can quickly process the constructed 3DLSG and command a robotic platform to visit a semantic-of-interest. The semantic-query take the form of: $``$Visit $v^F$ in $v^L$ of $v^T$ $"$, where  $v^F,v^L,v^T$ correspond to the semantic labels of the registered nodes in the final 3DLSG. The nodes are subsequently passed onto the hierarchical planner as the final terminal nodes  $v_{trm}^F$, $v_{trm}^L$ and $v_{trm}^T$ for further planning operation.

Figure~\ref{fig:HPP_resultsemQ_upd_collage} presents a response from the hierarchical path planner for a semantic query issued to the robot at the end of an inspection mission. Figure~\ref{fig:HPP_resultsemQ_upd_collage}(a)-(b) present the starting and terminal nodes of the query. The global landmark route generated by the planner is shown as a \textit{green} line over the Target-layer. The iteratively refined local layer routes (refer Fig~\ref{fig:HPP_resultsemQ_upd_collage}(c)-(d))) is shown as \textit{cyan} line over the \textit{Level} and \textit{Pose} layers respectively.

\section{Evaluation Setup}\label{sec:eval_setup}

\subsection{Hardware and Software}

\begin{figure}[htpb]
    \centering
    \includegraphics[width=\linewidth]{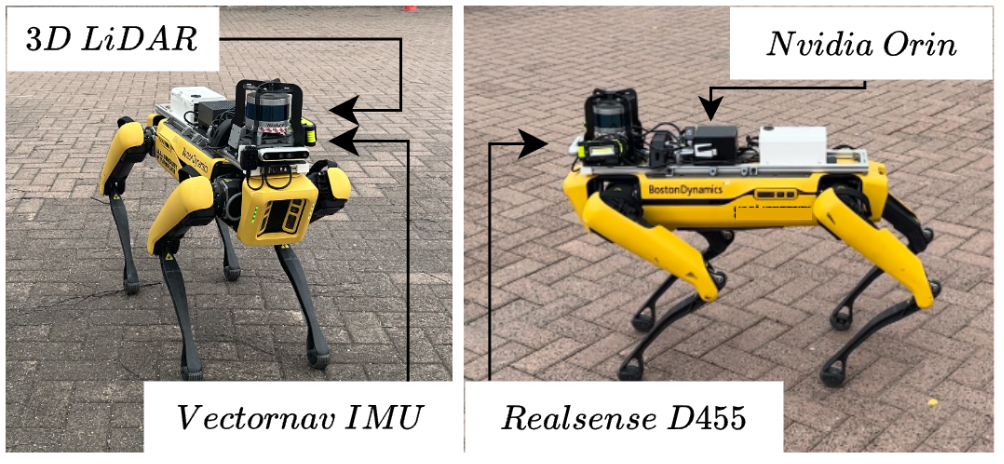}
    \caption{A visual depiction presenting the sensor suite deployed on a Boston Dynamics (BD) Spot quadruped robotic platform for deployment in the field.}
    \label{fig:hardware_setup}
\end{figure}

We validate the proposed architecture through field trials using a fully autonomous Boston Dynamics Spot quadruped, equipped with an NVIDIA Jetson Orin computing unit, a RealSense D455 stereo camera, a Velodyne VLP-16 3D LiDAR, and a VectorNav VN-100 IMU. Localization estimates are provided by the LIO-SAM framework~\cite{liosam2020shan}. Figure~\ref{fig:hardware_setup} illustrates the deployed autonomy sensor suite. For reference pose tracking, we use a nonlinear Model Predictive Controller (nMPC)~\cite{lindqvist2020nonlinear}, which generates velocity commands to Spot’s onboard low-level controller. We offload collision-avoidance to Spot's onboard avoidance module.

Additionally, the xFLIE architecture is evaluated in simulation using Gazebo~\cite{koenig2004design} and the RotorS aerial simulator~\cite{Furrer2016}. A key feature of xFLIE is its platform-agnostic design, requiring only that the robot be holonomic. Although large-scale aerial field trials are not presented in this work, the architecture was tested on a simulated aerial platform and deployed without significant modification on Spot. The only platform-specific adjustment involves a modality parameter that informs the inspection planner of the platform’s capability for vertical inspection.

The software stack runs on ROS Noetic with Ubuntu 20.04, with the Jetson Orin flashed using JetPack SDK v5.1. Graph construction is implemented using the NetworkX~\cite{hagberg2008exploring} library in Python 3.8. A voxel-grid point cloud down-sampler, based on the PCL library~\cite{Rusu_ICRA2011_PCL}, is implemented in C++. GPU resources are used primarily for executing YOLOv8n-seg, YOLOv8m-seg, and LightGlue models, while the rest of the xFLIE stack runs on CPU. Table~\ref{tab:general_params_vehicles} and ~\ref{tab:general_params_buildings} present the general parameters used during the experimental and simulated trials for the urban inspection scenario of \{\textit{cars},\textit{trucks}\} and the simulated trials for \textit{collapsed buildings}.
\begin{table}[t]
\centering
\caption{General parameters used during experimental and simulated trials for the urban inspection scenario of \{\textit{cars},\textit{trucks}\}.}
\label{tab:general_params_vehicles}
\begin{tabular}{@{}lcccl@{}}
\toprule
Symbol & Experimental & Simulated & Unit \\
\midrule
$\alpha$ & 86 & 69.4 & \si{\degree} \\
$d_{\max}$ & 30 & 30 & \si{\meter} \\
$I_w$ & 640 & 640 & \si{\pixel} \\
$I_h$ & 480 & 480 & \si{\pixel} \\
$d^{I}_{\text{thresh}}$ & 5 & 5 & \si{\meter} \\
$d^T_{check}$ & 4 & 4 & \si{\meter} \\
$d^F_{check}$ & 1.5 & 1.5 & \si{\meter} \\
$\Gamma_{\text{thresh}}$ & 0.3 & 0.3 & -- \\
$N$ & 3 & 3 & -- \\
$\epsilon$ & 2 & 2 & \si{\meter} \\
\bottomrule
\end{tabular}
\end{table}

\begin{table}[t]
\centering
\caption{General parameters used during simulated trials for the infrastructure inspection scenario of \textit{collapsed buildings}.}
\label{tab:general_params_buildings}
\begin{tabular}{@{}lcccl@{}}
\toprule
Symbol & Simulated & Unit \\
\midrule
$\alpha$ & 69.4 & \si{\degree} \\
$d_{\max}$ & 30  &\si{\meter} \\
$I_w$ & 640  &\si{\pixel} \\
$I_h$ & 480 & \si{\pixel} \\
$d^{I}_{\text{thresh}}$ & 5 & \si{\meter} \\
$d^T_{check}$ & 8 & \si{\meter} \\
$d^F_{check}$ & 1.5 & \si{\meter} \\
$\Gamma_{\text{thresh}}$ & 0.4 & -- \\
$N$ & 3 & -- \\
$\epsilon$ & 5 & \si{\meter} \\
\bottomrule
\end{tabular}
\end{table}
\subsection{Simulation Scenario}
\begin{figure}
    \centering
    \includegraphics[width=0.8\linewidth]{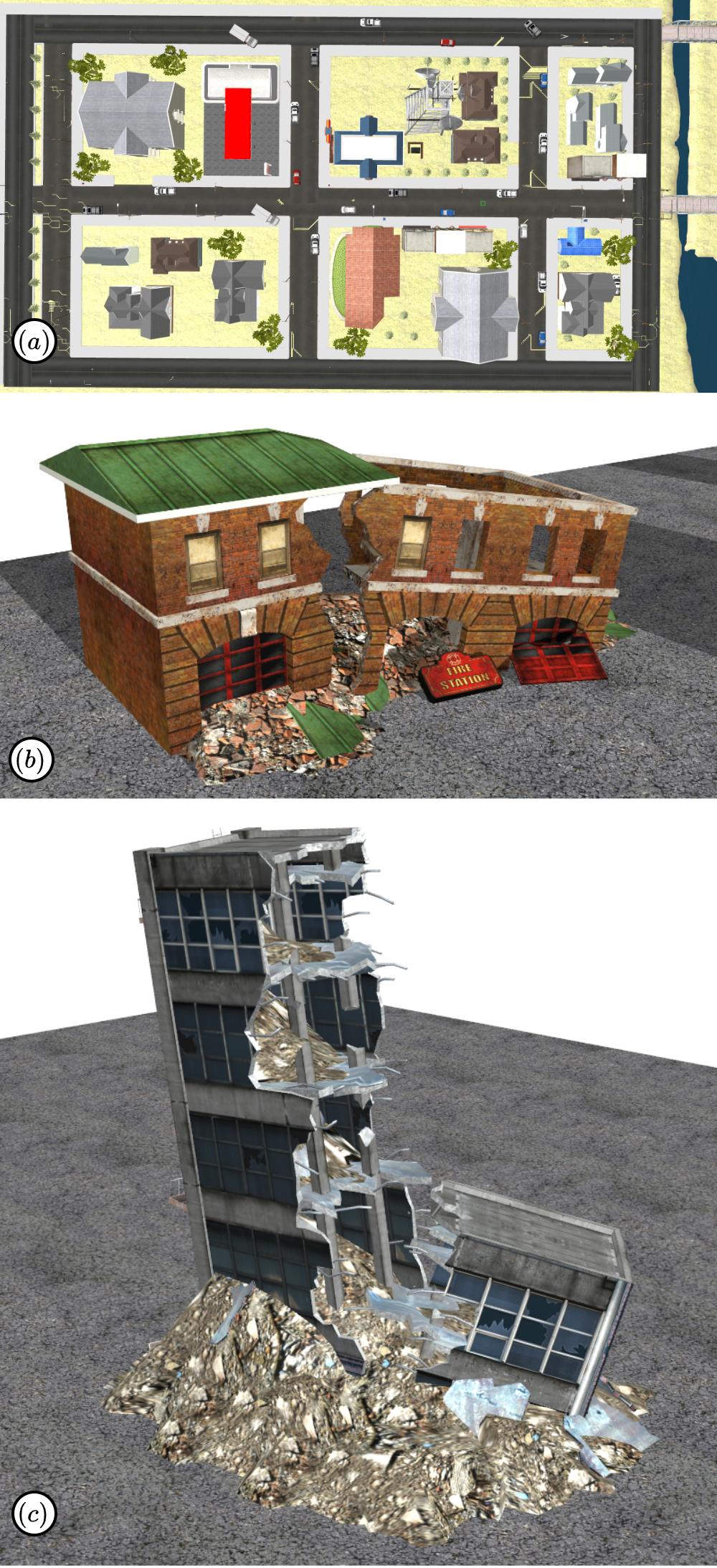}
    \caption{An overview of the simulated environment used to evaluate the proposed methodology.}
    \label{fig:sim_env_top-down}
\end{figure}

Figure.~\ref{fig:sim_env_top-down}(a) presents the simulated open-source city-scale environment (150 m [L] x 100 m [W] x 5 m [H]) used for the evaluation of xFLIE. The environment comprises of 20 vehicles, a combination between cars and trucks spread throughout the city. The proposed method is evaluated under three scenarios in the city-scale environment: (i) Scenario 1 simulates an inspection involving five vehicles (small-scale), (ii) Scenario 2 simulates an inspection involving ten vehicles (medium-scale), and (iii) Scenario 3 simulates an inspection involving twenty vehicles (large-scale). We utilized YOLO~\cite{yolov8_ultralytics} in this environment, specifically YOLOv8n-seg for the segmentation of \{cars,trucks\} semantics and a YOLOv8m-seg trained on Carparts dataset~\cite{car-seg-un1pm_dataset}. A demonstration video of the city-scale simulation, featuring the full inspection mission, hierarchical and semantic path‐planning for both system‐generated and operator‐defined queries is available at \url{https://youtu.be/cG1AHuhSHNc}.

Figure.~\ref{fig:sim_env_top-down}(b-c) present the solitary infrastructure inspection use-case for collapsed buildings environment. The proposed method is evaluated under two scenarios in this simulation: (1) Scenario 1 simulates the inspection of a collapsed fire-station and (ii) Scenario 2 simulates a collapsed industrial building. We implement CLIPSeg~\cite{luddecke2022image} to enable segmentation of target semantics \{\textit{buildings}\} and semantic features \{\textit{window}, \textit{person}, \textit{rubble}\}.

\subsection{Experimental Scenario}\label{sec:exp_setup}

\begin{figure}[htbp]
    \centering
    \includegraphics[width=\linewidth]{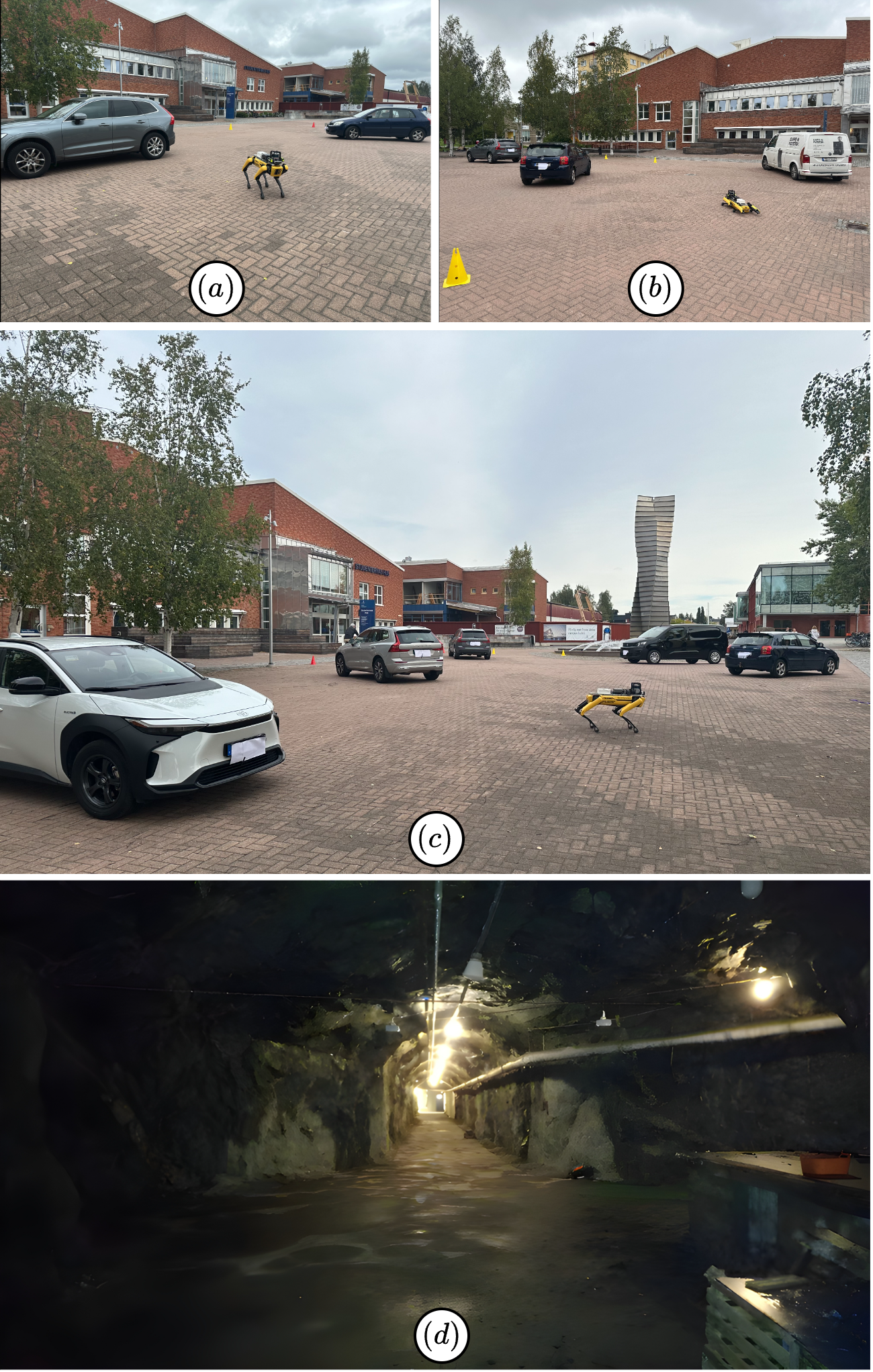}
    \caption{A visual description of the operating environment considered for field evaluation of xFLIE architecture.}
    \label{fig:exp_scene}
\end{figure}

Figure.~\ref{fig:exp_scene} presents the site of the field evaluations carried out in this work. We deploy the robot for inspecting vehicles in a campus environment (40 m [L] x 24 m [W] x 2 m [H]) (see Fig.~\ref{fig:exp_scene}(a-c)). We also present evaluation in a subterranean environment (10 m [L] x 6 m [W] x 4 m [H]) (see Fig.~\ref{fig:exp_scene}(d)).

We emphasize that the environments were not controlled. In the campus environment, though the vehicles designated for inspection were static, the scene in general was active with pedestrians, moving service vehicles and changing lighting conditions throughout the duration of field trials. The GPS-denied subterranean environment presented unstructured and low-light setup, allowing testing the autonomy stack in challenging conditions.
\begin{figure*}[htbp]
    \centering
    \includegraphics[width=0.9\linewidth]{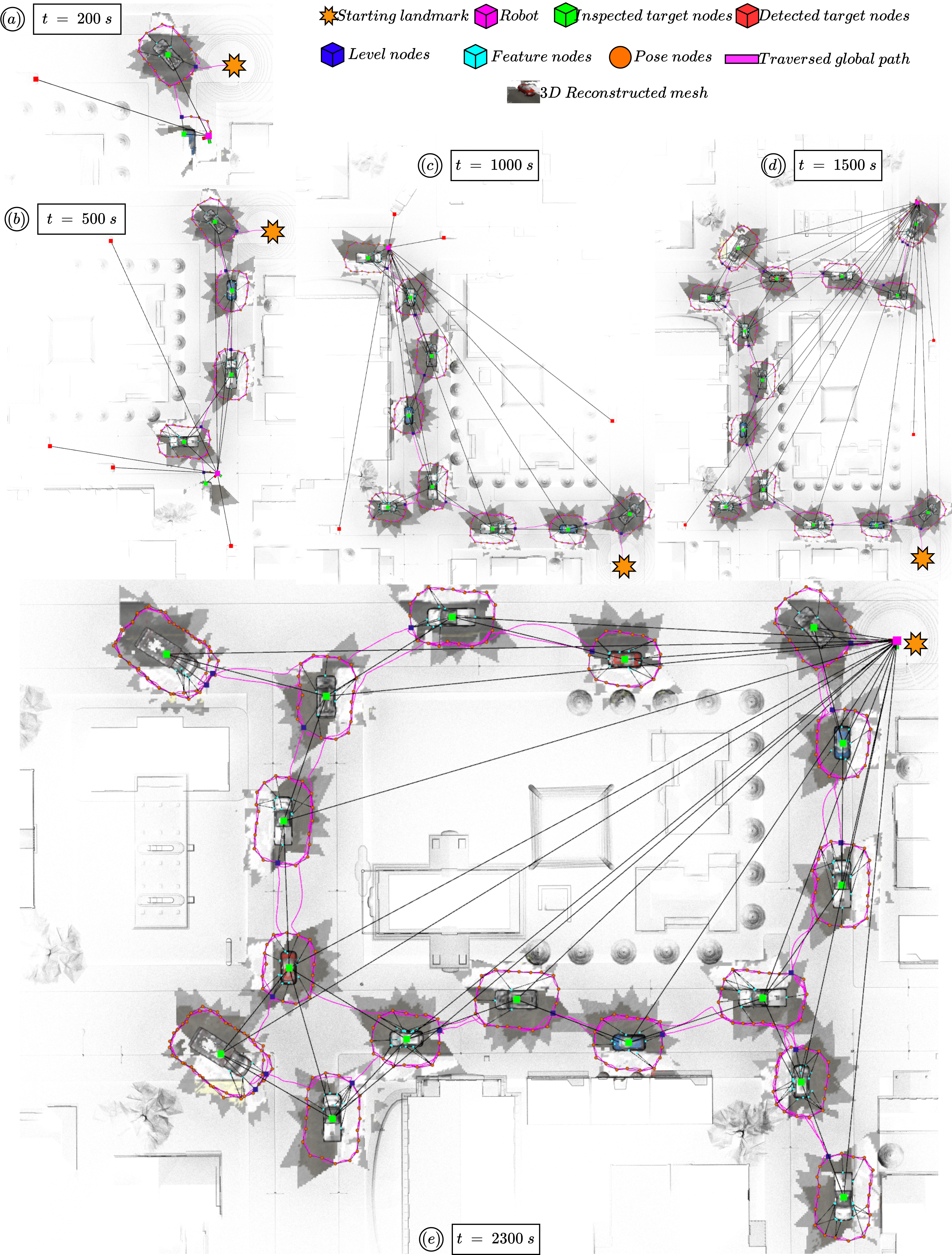}
    \caption{A collage of runtime snapshots of the proposed architecture evaluated in the simulated city-scale environment. (a)-(d) present the constructed 3DLSG during inspection over specified epochs of 200, 500, 1000 and 1700 seconds into the mission, respectively.(e) presents the overall 3DLSG constructed at the end of the simulated run.}
    \label{fig:run5_sim_collage}
\end{figure*}
A demonstration video of the field experiments is available online, \url{https://youtu.be/ODhKK8hr50o}. In this video, we visualize the performance of xFLIE architecture for the outdoor campus scenarios. We highlight the construction and utilization of the 3DLSG during autonomous inspection. We also demonstrate hierarchical and semantic path planning operations.

\section{Results and Discussions}\label{sec:res_des}

\subsection{Simulation Evaluation} \label{sec:sim_eval}
Figure~\ref{fig:run5_sim_collage} presents a sequence of runtime snapshots showing the evolution of the 3DLSG during different stages of the inspection mission. In this simulation run, the utility heuristic weights were set to $S_p=50$, $S_a=5$, and $S_n=5$ in Eq.~\eqref{eqn:utility_heuristic}. The framework successfully inspected 17 out of 20 available targets, achieving an 85\% inspection rate. Additional runs (see Table~\ref{tab:graph_population_stats_sims}) demonstrated improved performance, with detection and inspection of 19 targets (95\%).

Figures~\ref{fig:run5_sim_collage}(a)–(d) illustrate the progressive construction of the 3DLSG across its abstraction layers. Early in the mission (Fig.~\ref{fig:run5_sim_collage}(a)), the spatial proximity gain dominates the utility function, driving the robot to inspect nearby targets. As more semantic targets are detected and registered (Figs.~\ref{fig:run5_sim_collage}(b–c)), the influence of node centrality increases, leading the robot to prioritize clusters of targets. The final graph, shown in Fig.~\ref{fig:run5_sim_collage}(e), represents the fully constructed 3DLSG used for semantic query execution.

\begin{figure*}[hb]
    \centering
    \includegraphics[width=\textwidth]{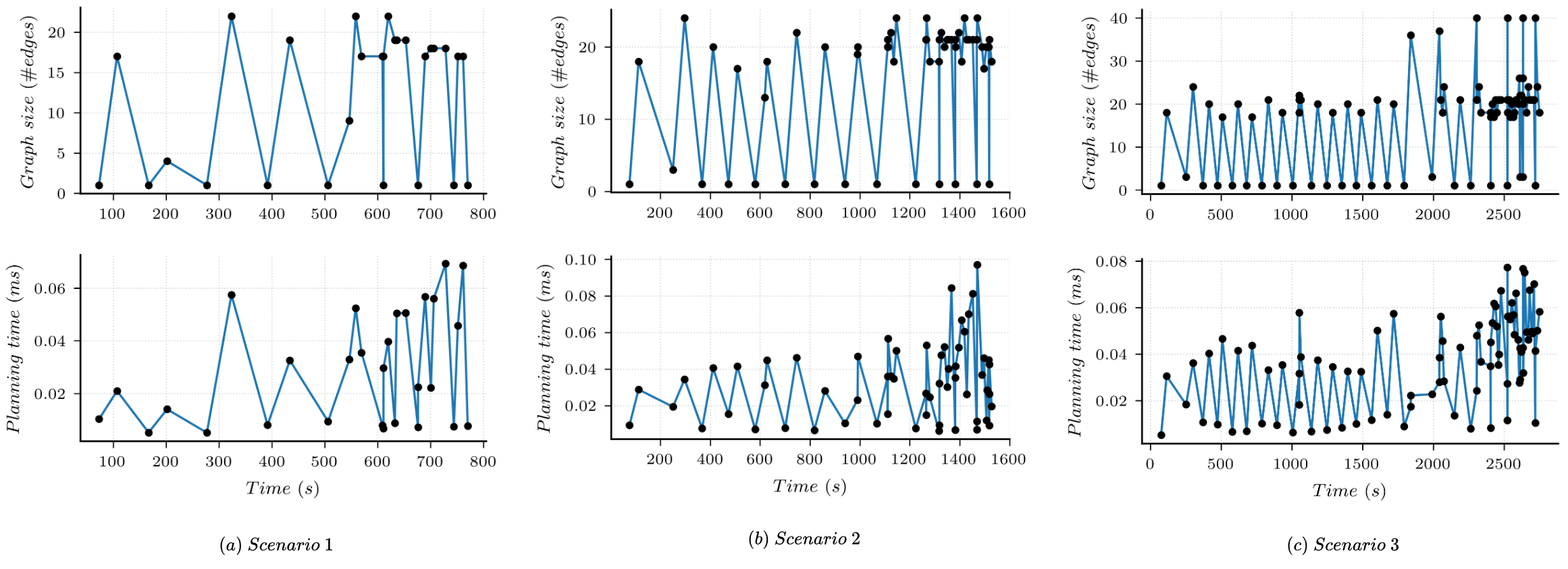}
    \caption{Quantitative assessment of planning performance recorded with respect to path computation time over input graph size for the city-scale simulation scenarios.}
    \label{fig:planning_performance_sims}
\end{figure*}

Figure~\ref{fig:planning_performance_sims} presents the pairwise relationship between planning time and graph size for each planning instance in the simulated city-scale environment. The results illustrate the influence of the 3DLSG’s nested structure on hierarchical path planning. Specifically, the planner expands only the relevant local subgraph associated with each \textit{Target}-layer node, effectively bounding the size of the graph processed per query. As shown in the figure, most planning instances operate on a median graph size of approximately 20 nodes, corresponding to the local $\mathcal{G}^P_k$ population. Instances with a graph size of 1 reflect queries resolved solely within the \textit{Level} layer, indicating that a single inspection level was sufficient for most semantic targets.

\begin{table*}[t]
  \centering
  \scriptsize
  \caption{Performance summary of xFLIE for the simulated scenarios.}
  \label{tab:graph_population_stats_sims}
  \resizebox{\textwidth}{!}{%
  \begin{tabular}{ll
                  *{5}{c}
                  *{2}{c}
                  c
                  c}
    \toprule
    \textbf{Environment} 
      & \textbf{Scenario} 
      & \multicolumn{5}{c}{\textbf{Cumulative Local Graph Population (Nodes/Edges)}} 
      & \multicolumn{2}{c}{\textbf{Unified Graph Population}} 
      & \shortstack[c]{\textbf{Targets}\\\textbf{Detected / Inspected}} 
      & \shortstack[c]{\textbf{Avg.\ Planning}\\\textbf{Time (ms)}} \\
    \cmidrule(lr){3-7}
    \cmidrule(lr){8-9}
      & 
      & \textbf{Target} 
      & \textbf{Level} 
      & \textbf{Pose} 
      & \textbf{Feature} 
      & \textbf{Total} 
      & \textbf{Nodes/Edges} 
      & \textbf{Reconciled (\%)} 
      &  &  \\
    \midrule
    \multirow{3}{*}{\textbf{City-scale (Vehicles)}} 
      & Scenario 1  
        &  6/9   & 10/5   & 106/101  & 68/36   & 190/151  
        & 138/146  & 27.3
        & 100.0/100.0 & 0.0287 \\
      & Scenario 2  
        & 11/21  & 21/11  & 253/244  & 102/55  & 387/332  
        & 298/310  & 22.3 
        & 100.0/100.0 & 0.0327 \\
      & Scenario 3  
        & 20/40  & 40/23  & 469/452  & 219/121 & 748/636  
        & 572/581  & 23.5  
        &  95.0/100.0 & 0.0371 \\
    \addlinespace
    \multirow{2}{*}{\textbf{Collapsed Buildings}} 
      & Scenario 1  
        &  2/1   & 4/5   & 84/162  &  93/50  & 183/218  
        & 124/218  & 32.2 
        &  100.0/100.0 & 0.0213 \\
      & Scenario 2  
        & 2/1   & 10/17  & 174/330  & 91/47  & 277/395  
        & 213/395  & 23.0 
        &  100.0/100.0 & 0.0551 \\
    \bottomrule
  \end{tabular}%
  }
\end{table*}

Table~\ref{tab:graph_population_stats_sims} summarizes the performance of the proposed framework across simulation scenarios, reporting 3DLSG construction statistics, inspection success rates, and planning efficiency. The table highlights the cumulative growth of each abstraction layer and overall graph population. We also report detection and inspection success rates, along with average planning times for hierarchical and semantic queries.

Across all scenarios, the 3DLSG was constructed as expected for detected inspection targets. In both city-scale and collapsed-building environments, the full pipeline from segmentation and localization to graph construction was validated in most cases. In Scenario 3 (city-scale), a reduction in detection success was observed due to one missed segmentation, likely caused by perceptual uncertainty from distant viewpoints during exploration. This led to an 85\% target detection rate in one run, though the framework maintained a 100\% inspection rate for all detected targets.

The hierarchical planner scaled effectively with graph size. In the city-scale scenario, it handled nearly a twofold increase in graph nodes with only a 12\% increase in average planning time. In the collapsed-building environment, the planner maintained an average planning time of 0.04 ms. The increased \textit{Level} layer population reflects multi-level inspection behaviour, while denser populations in the \textit{Pose} and \textit{Feature} layers are consistent with view-poses and feature observations accumulated during inspection.

\subsection{Field Evaluation} \label{sec:exp_eval}

\begin{figure*}[htbp]
    \centering
    \includegraphics[width=\linewidth]{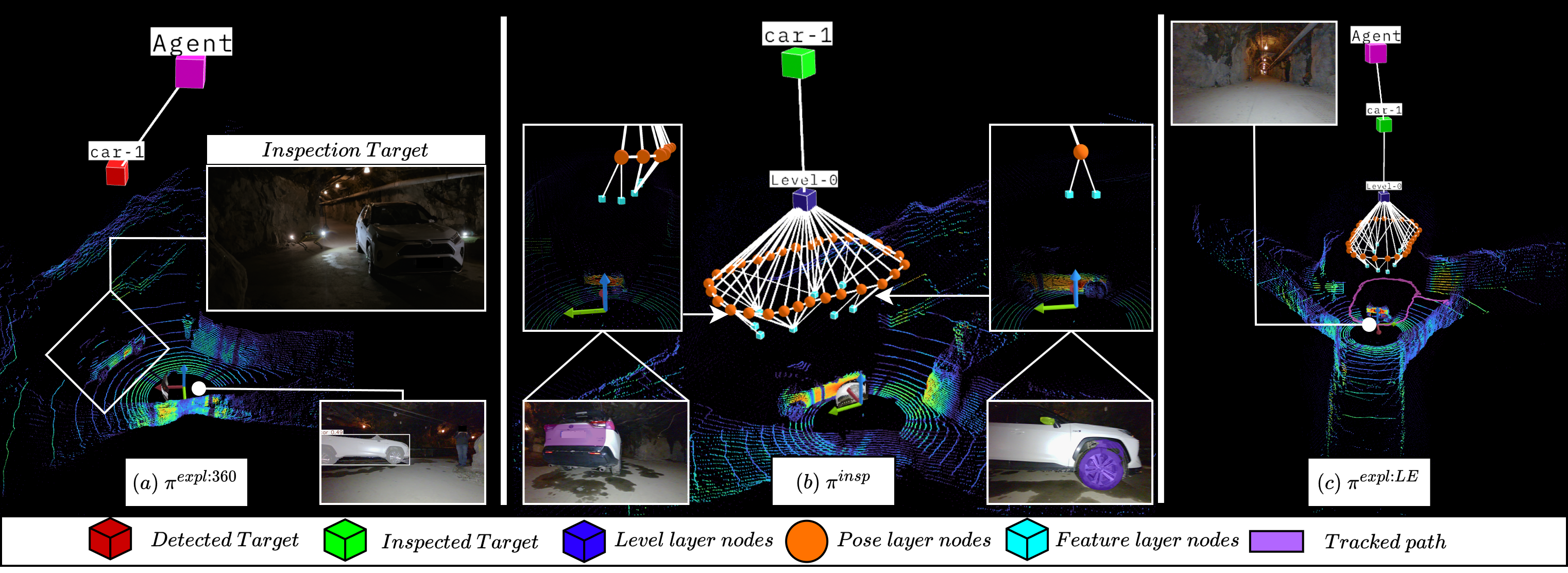}
    \caption{Qualitative results from inspection experiment in a subterranean environment.(a) showcases the detection and registration of a semantic target in the \textit{Target} layer graph, (b) presents the constructed 3DLSG during inspection and (c) captures the robot executing $\pi^{expl:LE}$ for detection of nearby semantic targets around the current vicinity.}
    \label{fig:run3_exp_subt}
\end{figure*}

Figure~\ref{fig:run3_exp_subt} presents qualitative results from the autonomous inspection of a \textit{car} semantic target in a subterranean environment. Due to spatial constraints, a single-vehicle inspection scenario is shown. The robot successfully detected, inspected, and constructed the 3DLSG representation during the mission.

Figure~\ref{fig:run3_exp_subt}(a) shows the registration of \textit{car-1} into $\mathcal{G}^T_k$, observed during the $\pi^{expl:360}$ behavior. The unified 3DLSG constructed during the inspection is illustrated in Fig.~\ref{fig:run3_exp_subt}(b), with snapshots of the incremental construction of the \textit{Pose} and \textit{Feature} layers shown to its left and right, respectively. Finally, Fig.~\ref{fig:run3_exp_subt}(c) depicts a local surveying instance post-inspection, including an inset RGB image from the robot’s point of view.

\begin{figure*}[htbp]
    \centering
    \includegraphics[width=\linewidth]{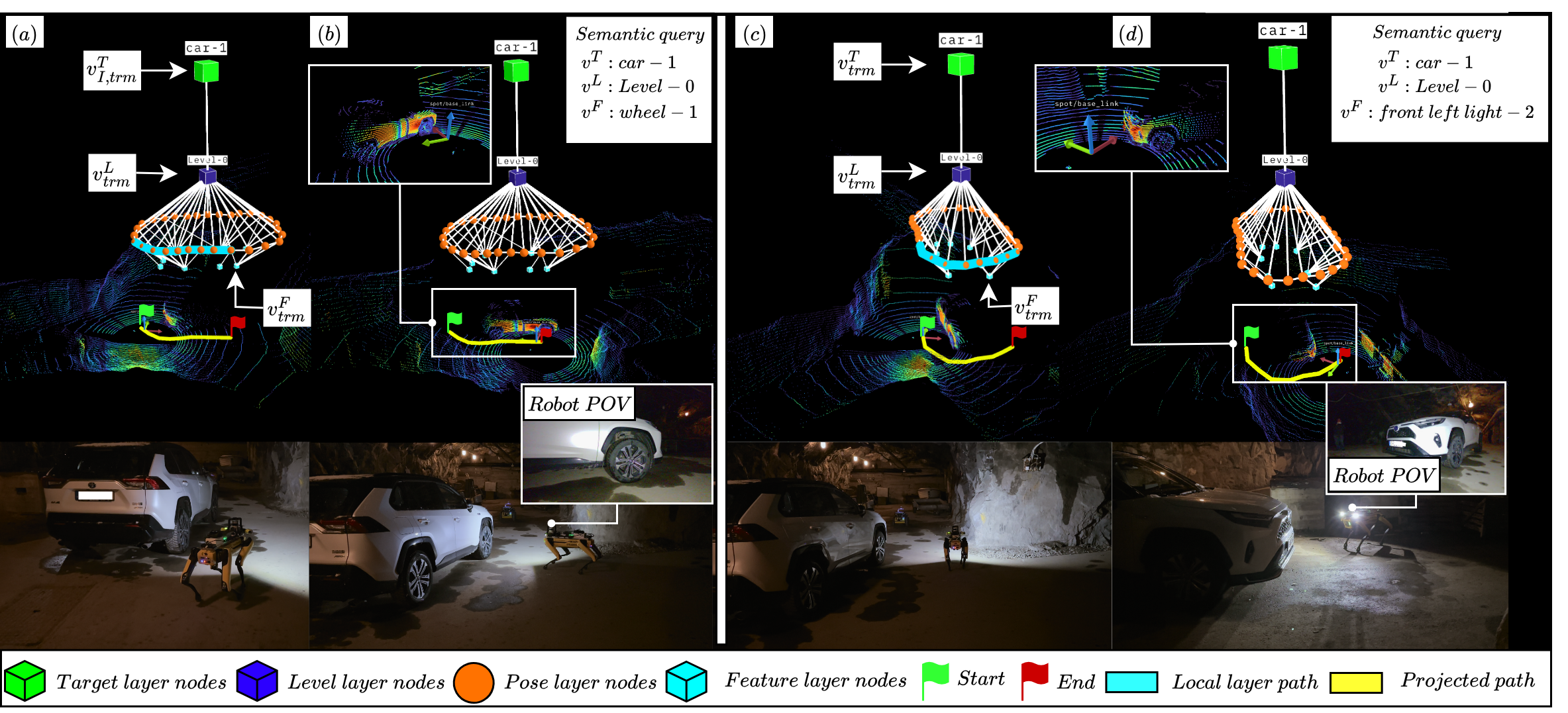}
    \caption{Qualitative results from semantic path-planning in subterranean environment experiments.}
    \label{fig:run3_subt_exp_spp}
\end{figure*}

Figure~\ref{fig:run3_subt_exp_spp} shows the robot navigating in response to operator-defined semantic queries. Figures~\ref{fig:run3_subt_exp_spp}(a–b) illustrate the system processing the query: \textit{“Observe wheel-1 in Level-0 of car-1”}, while Figs.~\ref{fig:run3_subt_exp_spp}(c–d) show results for the query: \textit{“Observe front-left-light-2 in Level-0 of car-1”}. In both cases, the framework successfully mapped the semantic labels to corresponding nodes in the 3DLSG and generated a valid path. The planned local-layer path is shown in solid \textit{cyan}, with its ground-projected counterpart in solid \textit{yellow}, along with the \textit{start} and \textit{end} positions of the commanded trajectory.

\begin{figure*}[htbp]
    \centering
    \includegraphics[width=\linewidth]{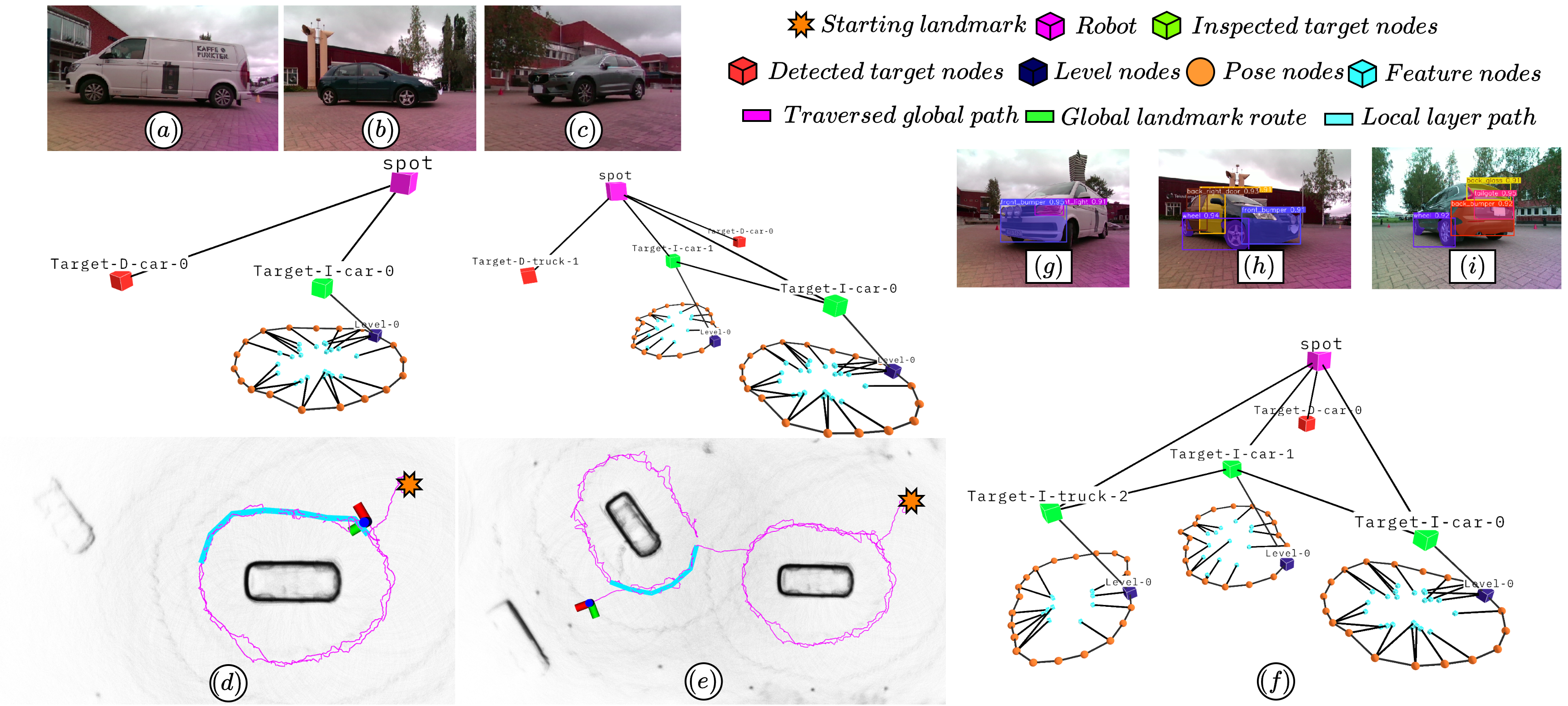}
    \caption{Visualization of experimental results obtained during three vehicle inspection scenario. (a)-(c) presents the different vehicles located in the operating environment. (d) presents the outcome of hierarchical planning in response to a planning query to navigate towards $\textit{Target-D-car-0}$ for further inspection. In (e), another instance of leveraging the 3DLSG by the FLIE autonomy for path-planning tasks is shown, (f) presents the 3DLSG constructed during the evaluated inspection mission. (g)-(i) shows snapshots of the performance of the onboard segmentation model to detect and extract desired semantic features observed during inspection of semantic targets observed in the environment.}
    \label{fig:threecar_exp_v1_collage}
\end{figure*}

Figure~\ref{fig:threecar_exp_v1_collage} presents a compilation of experimental results from the inspection of three cars, along with the final constructed 3DLSG. The system successfully detected, localized, and inspected all semantic targets in the environment. Figures~\ref{fig:threecar_exp_v1_collage}(a–c) show the target vehicles designated for inspection, while Figs.~\ref{fig:threecar_exp_v1_collage}(d–e) illustrate the performance of the hierarchical planner responding to FLIE-generated queries over the nested 3DLSG. The final scene graph is visualized in Fig.~\ref{fig:threecar_exp_v1_collage}(f). An anomalous node $v_D^T$ is present, caused by a semantic localization mismatch where a target was falsely localized in a region without vehicles. The mission was terminated after completing inspection of all three vehicles. Figures~\ref{fig:threecar_exp_v1_collage}(g–i) show example segmentations of observed semantic features registered into $\Tilde{\mathcal{G}}_{PF}$ during inspection.

\begin{figure*}[htbp]
    \centering
    \includegraphics[width=\linewidth]{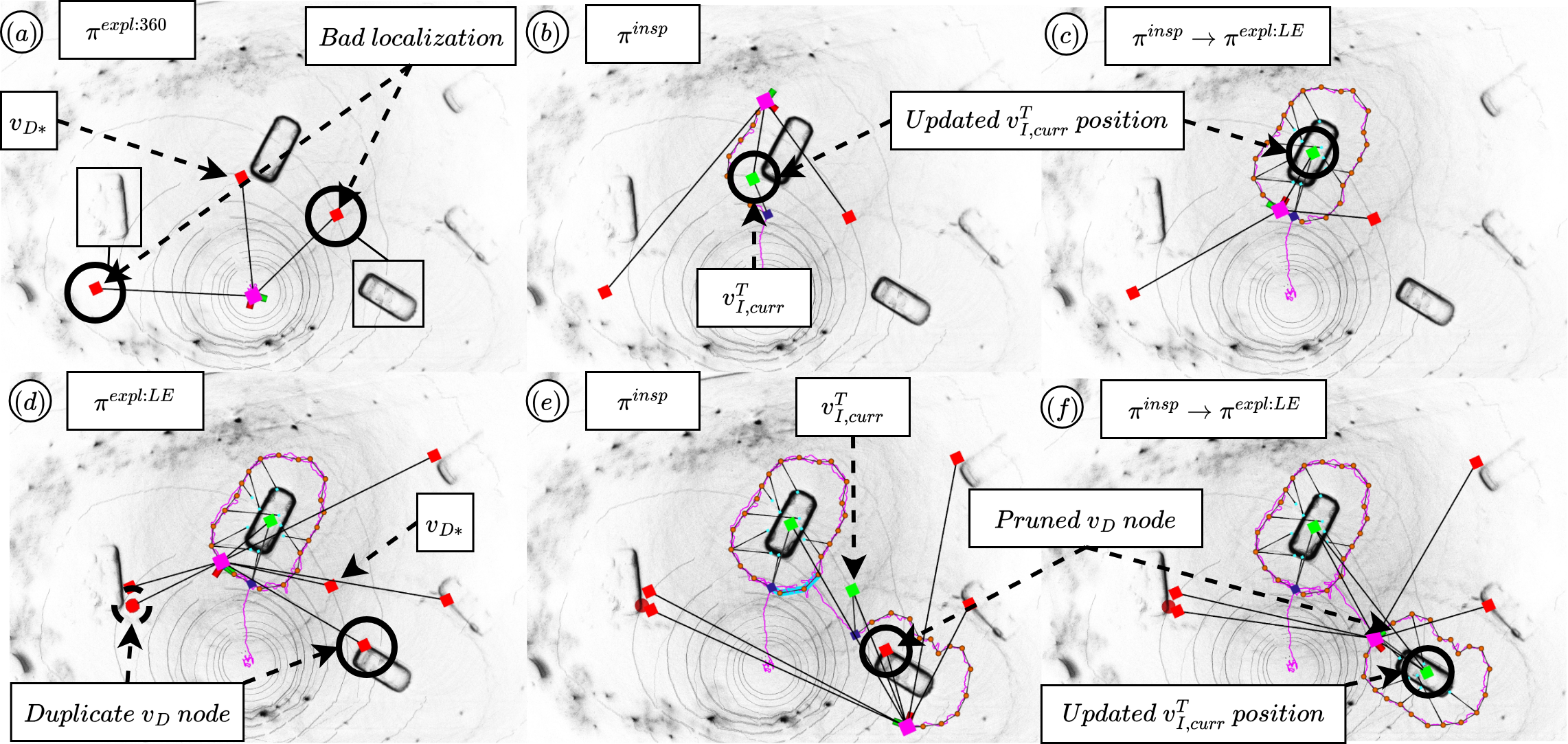}
    \caption{A visual collage highlighting key moments during a five-vehicle inspection mission, where the $\textit{Target}$ layer graph is pruned to remove low-quality and redundant semantic target observations.}
    \label{fig:run4_GToptim_sequence}
\end{figure*}

The influence of the node-centrality gain $S_n$ is evident in this scenario (refer to the evaluation video in Sections~\ref{sec:eval_setup}–\ref{sec:exp_setup}). After inspecting \textit{Target-I-car-0}, the robot was directed to the distant \textit{Target-I-car-2} due to the presence of duplicate target nodes in close proximity, which inflated the node degree $N(v_D)$ for that region (see middle-left of Fig.~\ref{fig:run4_GToptim_sequence}(e)). Although the implemented optimization removed most redundant registrations, it failed to eliminate this specific outlier, which influenced the planning decision and led the robot to skip a nearer target. This highlights a residual limitation in handling edge-case redundancies in $\mathcal{G}^T_k$.

Figure~\ref{fig:run4_GToptim_sequence} illustrates a sequence of runtime instances where the constructed $\mathcal{G}T$ is optimized and pruned during a five-vehicle inspection experiment. The initial graph, built at the end of the $\pi^{expl:360}$ behaviour, is shown in Fig.~\ref{fig:run4_GToptim_sequence}(a). A synchronization issue between YOLO-based RGB processing and the robot’s odometry estimate $\bm{x}^p_k$ results in a misaligned semantic localization, highlighted by the bold dotted circle. This de-synchronization caused by continuous pose updates during $\pi^{expl:360}$ leads to an incorrect estimate $\bm{p}(v_D)$, which is later corrected in the final graph update.

Figures~\ref{fig:run4_GToptim_sequence}(b–c) show the timeline of inspecting a semantic target $v_{D*}$. The initially estimated pose (b) is refined after inspection by aligning $\bm{p}(v_D)$ with the centroid of the reconstructed point cluster $\texttt{P}$ around $v_T^{I,curr}$ (c). Figures~\ref{fig:run4_GToptim_sequence}(d–e) depict the node validation process: new targets observed during $\pi^{expl:LE}$ are added to $\mathcal{G}^T_k$ (d), followed by execution of $\pi^{insp}$ (e). Finally, Fig.~\ref{fig:run4_GToptim_sequence}(f) shows an example of pruning a duplicate node via a polygon-based containment check removing a registration made during exploration that falls within a restricted region around an already inspected target.

\begin{figure*}[htp]
    \centering
    \includegraphics[width=\textwidth]{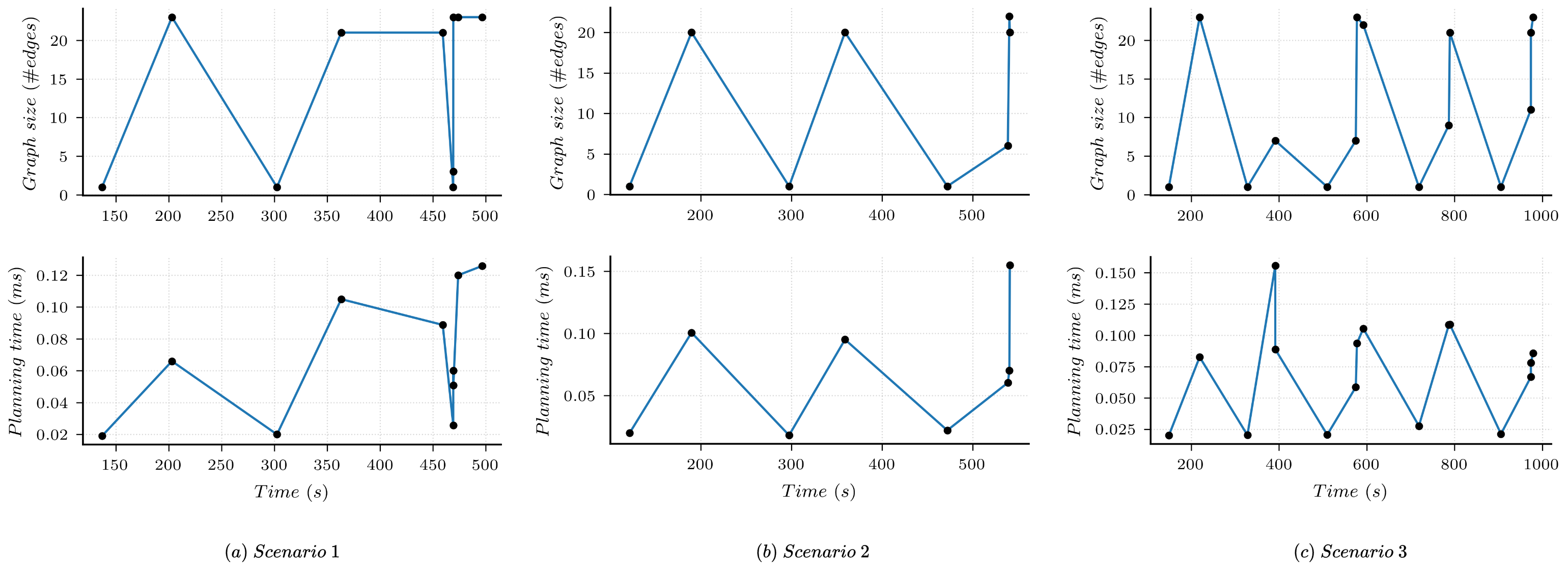}
    \caption{Assessment of path-planning performance recorded with respect to path computation time over input graph size for outdoor campus experiments.}
    \label{fig:planning_performance_exp}
\end{figure*}

Figure~\ref{fig:planning_performance_exp} shows the evolution of the 3DLSG and corresponding planning times across the experimental scenarios. The framework successfully detected and inspected all available targets in both environments, with abstraction layers populated as expected. This is reflected in the dense growth of the \textit{Pose} and \textit{Feature} layers over time. As target density and spatial distribution increased, the system maintained robust connectivity by establishing traversable edges between inspected targets.

Table~\ref{tab:graph_population_stats_exp} summarizes the graph construction and planning performance across outdoor campus and subterranean environments. On average, the iterative merging of abstraction layers resulted in a 26\% reduction in cumulative node count within the unified 3DLSG. The high edge count in the \textit{Pose} layer is due to symbolic parent-child connections introduced during inspection (see Fig.~\ref{fig:run3_exp_subt}).

Planning time showed a marginal increase across scenarios, indicating good scalability. In Scenarios 1 and 2 (outdoor campus), the \textit{Feature} layer exhibited a higher node density compared to the \textit{Pose} layer, attributed to a greater number of features observed per view-pose during inspection.

\begin{table*}[htbp]
  \centering
  \scriptsize
  \caption{Performance summary of xFLIE for the experimental scenarios.}
  \label{tab:graph_population_stats_exp}
  \resizebox{\textwidth}{!}{%
  \begin{tabular}{ll
                  *{5}{@{\quad}c@{\quad}}
                  *{2}{@{\quad}c@{\quad}}
                  c
                  c}
    \toprule
    \textbf{Environment} 
      & \textbf{Scenario} 
      & \multicolumn{5}{c}{\textbf{Cumulative Local Graph Population (Nodes/Edges)}} 
      & \multicolumn{2}{c}{\textbf{Unified Graph Population}} 
      & \shortstack[c]{\textbf{Targets}\\\textbf{Detected / Inspected (\%)}} 
      & \shortstack[c]{\textbf{Avg.\ Planning}\\\textbf{Time (ms)}} \\
    \cmidrule(lr){3-7}
    \cmidrule(lr){8-9}
    \cmidrule(lr){10-11}
      & 
      & \textbf{Target} 
      & \textbf{Level} 
      & \textbf{Pose} 
      & \textbf{Feature} 
      & \textbf{Total} 
      & \textbf{Nodes/Edges} 
      & \textbf{Reconciled (\%)} 
      &  &  \\
    \midrule

    \multirow{3}{*}{\textbf{Outdoor Campus (Vehicles)}} 
      & Scenario 1  
        &  3/3   & 4/2   & 48/46  & 59/36   & 114/87  
        & 83/85  & 27.1
        & 100.0/100.0 & 0.0681 \\
      & Scenario 2  
        & 5/6  & 6/3  & 68/65  & 80/47  & 159/121  
        & 114/118  & 28.3 
        & 100.0/100.0 & 0.0652 \\
      & Scenario 3  
        & 7/11  & 10/5  & 119/114  & 57/29 & 193/159  
        & 148/124  & 23.3  
        & 100.0/100.0 & 0.0707 \\

    \addlinespace
    \textbf{Subterranean (Vehicles)}  
      & Scenario 1  
        &  2/1   & 4/5   & 84/162  & 93/50   & 183/218  
        & 124/218  & 32.2 
        & 100.0/100.0 & 0.0523 \\
    \bottomrule
  \end{tabular}%
  }
\end{table*}

\subsection{Feasibility Analysis of Planning Performance}\label{sec:trad_off}

We evaluate the efficiency of planning over hierarchical graphs compared to conventional volumetric-based approaches in the context of an autonomous inspection mission, where a robot must compute a path to a specified semantic target. The primary metric is planning time, measured as the environment scales. We benchmark against $D^*_+$~\cite{karlsson2022d+}, a publicly available, grid-based planner that operates on volumetric occupancy maps such as Octomap~\cite{wurm2010octomap}.

Planning performance is recorded for each scenario within a simulated city-scale environment. We evaluate the performance for each scenario We compare the hierarchical planner (HP) against the volumetric planner (VP) across two map resolutions: $0.5\si{m}$ ($\mathbb{M}_{0.5}$) and $0.7\si{m}$ ($\mathbb{M}_{0.7}$). The remaining parameters are kept at default values: a risk parameter of 0.2 and a spline step size of 0.166.

\begin{table*}[t]
  \centering
  \scriptsize
  \setlength{\tabcolsep}{4pt}
  \renewcommand{\arraystretch}{1.1}
  \caption{Analysis of planning performance during simulated scenarios in the city-scale environment.}
  \label{tab:planning_results}
  \resizebox{\textwidth}{!}{%
    \begin{tabular*}{\textwidth}{@{\extracolsep{\fill}}
      >{\raggedright\arraybackslash}p{5.2cm}
      S[table-format=1.2e-1] S[table-format=3.2]
      S[table-format=1.2e-1] S[table-format=3.2]
      S[table-format=1.3]    S[table-format=3.3]
      S[table-format=1.3]    S[table-format=3.3]
    }
    \toprule
    & \multicolumn{4}{c}{\textbf{Hierarchical Path Planner}}
    & \multicolumn{4}{c}{\textbf{VP Planner}} \\
    \cmidrule(lr){2-5} \cmidrule(lr){6-9}
    \multicolumn{1}{l}{\textbf{Query}}
      & \multicolumn{2}{c}{\textbf{Target layer}}
      & \multicolumn{2}{c}{\textbf{Pose layer}}
      & \multicolumn{2}{c}{$\mathbb{M}_{0.5}$}
      & \multicolumn{2}{c}{$\mathbb{M}_{0.7}$} \\
    \cmidrule(lr){2-3} \cmidrule(lr){4-5} \cmidrule(lr){6-7} \cmidrule(lr){8-9}
    & \multicolumn{1}{c}{\shortstack[c]{Planning\\Time (\si{s})}}
    & \multicolumn{1}{c}{\shortstack[c]{Path\\Length (\si{m})}}
    & \multicolumn{1}{c}{\shortstack[c]{Planning\\Time (\si{s})}}
    & \multicolumn{1}{c}{\shortstack[c]{Path\\Length (\si{m})}}
    & \multicolumn{1}{c}{\shortstack[c]{Planning\\Time (\si{s})}}
    & \multicolumn{1}{c}{\shortstack[c]{Path\\Length (\si{m})}}
    & \multicolumn{1}{c}{\shortstack[c]{Planning\\Time (\si{s})}}
    & \multicolumn{1}{c}{\shortstack[c]{Path\\Length (\si{m})}} \\
    \midrule

    \multicolumn{9}{l}{\textbf{Scenario 1}} \\
    \midrule
    Visit \textit{front-bumper-1} in \textit{Level-0} of \textit{truck-4}
      & \num{1.7e-5}  & \num{32.33}
      & \num{1.1e-4}  & \num{49.92}
      & \num{0.526}   & \num{45.420}
      & \num{0.220}   & \num{48.600} \\

    Visit \textit{back-left-door-1} in \textit{Level-0} of \textit{car-0}
      & \num{2.2e-5}  & \num{48.44}
      & \num{1.1e-4}  & \num{56.86}
      & \num{0.632}   & \num{45.229}
      & \num{0.250}   & \num{46.536} \\

    Visit \textit{front-bumper-1} in \textit{Level-0} of \textit{car-1}
      & \num{1.6e-5}  & \num{16.11}
      & \num{5.8e-5}  & \num{19.74}
      & \num{0.560}   & \num{16.020}
      & \num{0.110}   & \num{17.180} \\

    \addlinespace
    \midrule
    \multicolumn{9}{l}{\textbf{Scenario 2}} \\
    \midrule
    Visit \textit{trunk-1} in \textit{Level-0} of \textit{truck-9}
      & \num{4.0e-5}  & \num{49.80}
      & \num{2.1e-4}  & \num{70.83}
      & \num{1.085}   & \num{57.479}
      & \num{0.431}   & \num{59.580} \\

    Visit \textit{front-glass-1} in \textit{Level-0} of \textit{truck-6}
      & \num{3.5e-5}  & \num{89.44}
      & \num{3.2e-4}  & \num{105.86}
      & \num{1.150}   & \num{76.000}
      & \num{0.369}   & \num{77.179} \\

    Visit \textit{front-bumper-1} in \textit{Level-0} of \textit{truck-2}
      & \num{2.7e-5}  & \num{68.60}
      & \num{2.2e-4}  & \num{66.53}
      & \num{1.030}   & \num{53.414}
      & \num{0.390}   & \num{53.815} \\

    \addlinespace
    \midrule
    \multicolumn{9}{l}{\textbf{Scenario 3}} \\
    \midrule
    Visit \textit{front-bumper-1} in \textit{Level-0} of \textit{car-9}
      & \num{7.74e-5} & \num{85.16}
      & \num{3.4e-4}  & \num{95.85}
      & \num{1.7}     & \num{77.63}
      & \num{0.75}    & \num{80.76} \\

    Visit \textit{front-glass-1} in \textit{Level-1} of \textit{truck-16}
      & \num{7.71e-5} & \num{136.48}
      & \num{5.4e-4}  & \num{145.50}
      & \num{1.9}     & \num{102.15}
      & \num{0.53}    & \num{103.40} \\

    Visit \textit{front-right-door-1} in \textit{Level-0} of \textit{truck-18}
      & \num{7.6e-5}  & \num{110.69}
      & \num{4.1e-4}  & \num{119.52}
      & \num{0.35}    & \num{73.08}
      & \num{0.45}    & \num{74.19} \\
    \bottomrule
    \end{tabular*}%
  }
\end{table*}

For each operator-defined semantic query, the system parses the query to identify the target terminal node that the robot must reach. This target is then provided to both planners: the hierarchical planner (HP) receives the corresponding node in the 3DLSG, while the volumetric planner (VP) is given the target's positional coordinates. HP initializes planning from the robot’s current location within the 3DLSG, whereas VP uses the current odometry estimate as the starting point. The HP output is visualized in three components: (a) the high-level global route over the \textit{Target} layer (solid \textit{green}), (b) the traversable local-layer path (solid \textit{cyan}), and (c) its ground-projected representation (solid \textit{yellow}). The VP route is visualized in solid \textit{red}. We exclude traversal data for the \textit{Level} layer in Table~\ref{tab:planning_results}, as most city-scale scenarios involved inspection within a single level.

A summary of cumulative planning time and path length is presented in Table~\ref{tab:planning_results}, measured across the \textit{Target} and \textit{Pose} layers. Overall, the hierarchical planner (HP) achieved multiple orders of magnitude reduction in planning times compared to the volumetric planner (VP) at map resolutions of $\mathbb{M}_{0.5}$ and $\mathbb{M}_{0.7}$, respectively, across various scenarios. In terms of path length, HP produced, on average, 1.29x and 1.25x longer traversable routes to the goal for the corresponding resolutions. Although the computed paths are slightly longer, HP offers a substantial gain in computational efficiency. These findings underscore the scalability and effectiveness of hierarchical scene representations over conventional volumetric maps in the context of navigation during large-scale inspection missions.

\subsection{Comparison Study}

A direct comparison with equivalent 3DSG-based methods could not be made due to differences in the problem domains being addressed~\cite{Hughes2022HydraAR,bavle2022s,greve2024collaborative}. Instead, we compare the proposed method against the following:

(i) \textit{FLIE + 3DLSG + D$^{*+}$}: This configuration forms the \textit{baseline} for evaluation. The hierarchical path planner is replaced with the D$^{*+}$ method to simulate decision-making over the 3DLSG and navigation within a volumetric scene model. A map resolution of 0.4~$\si{\meter}$ ($\mathbb{M}_{0.4}$) is used.

(ii) \textit{GBPlanner2.0}: This is a graph-based exploration planner designed for mapping previously unknown environments. The planner operates over a topological graph extracted from the incrementally built geometric model of the environment. We adopt the default map resolution of 0.2~$\si{\meter}$ ($\mathbb{M}_{0.2}$).

We model the maximum sensing range $d_{max}$ as defined in Table~\ref{tab:general_params_vehicles} for all methods. Each method is evaluated within the city-scale environment. We run the scenarios with the robot starting in various locations in the environment. A mission time limit of 700, 1400 and 2800 seconds is observed for each scenario. This value corresponds to the average runtime of the proposed method to complete the mission. We run four copies of experiments for each scenario. Finally, we evaluate the performance across three main metrics provided as follows:

(i) \textit{Computational Load} (\%): This is computed from the CPU usage of each method, averaged out for the number of trials. Naturally, this metric aims to underscore the relevance of planning over a hierarchical scene model compared to volumetric methods for navigation of large-scale environments.

(ii) \textit{Inspection Coverage} ($\si{\meter}^3$): This refers to the proportion of the target’s enclosed volume represented by the bounding box generated from the segmented points obtained when the target was observed during inspection. We refer to the work in~\cite{saucedo2024box3d} to generate the bounding box around inspection targets.

(iii) \textit{Root Mean Square Error (RMSE)} ($\si{\meter}$): This is computed between the estimated and ground-truth locations of inspected semantic targets. For the proposed method, we consider the registered metric coordinates as positional information of the target nodes. For others, we consider the centroid of the bounding box generated around each target based on the segmented pointcloud information.

Table~\ref{tab:method_comparison} summarizes the results of the comparison study. In all three scenarios, the proposed method outperforms the baseline and the exploration planner in all three metrics. xFLIE maintains upto 52\%-59\% lower computational overhead compared to exploration planner and the baseline method in Scenario 1. Moreover, the proposed method demonstrates better target localization performance, achieving approximately 10\% lower RMSE than the baseline and about 74.24\% lower RMSE than the pure exploration planner during the mission. The higher accuracy by xFLIE is attributed to the consideration of the containment polygon (refer Section~\ref{sec:xFLIE}-\ref{sec:xflie_insp}), whose centroid is used to update the target node position after its inspection.

\begin{table}[t]
  \centering
  \scriptsize
  \setlength{\tabcolsep}{3pt}
  \renewcommand{\arraystretch}{1.05}
  \caption{Comparison of planning frameworks evaluated for simulated city-scale environment.}
  \label{tab:method_comparison}
  \scalebox{0.8}{%
    \begin{tabular}{
      >{\raggedright\arraybackslash}p{5.0cm}
      S[table-format=3.3]
      S[table-format=3.2]
      S[table-format=1.3]
    }
    \toprule
    \textbf{Scenario / Method}
      & \multicolumn{1}{c}{\shortstack[c]{Computational\\Load (\%)}}
      & \multicolumn{1}{c}{\shortstack[c]{Inspection\\Coverage ($\si{\meter}^3$)}}
      & \multicolumn{1}{c}{\shortstack[c]{Target Localization\\RMSE ($\si{\meter}$)}} \\
    \midrule

    \multicolumn{4}{l}{\textbf{Scenario 1}} \\
    \midrule
    \hspace{2ex}FLIE + Target layer graph + D$^{*+}$ (\textit{Baseline}) & 103.57  & 103.02 & 0.19 \\
    \hspace{2ex}GBPlanner2.0                                              & 87.63   & 64.19  & 0.66 \\
    \hspace{2ex}xFLIE (\textbf{Proposed})                                  & 42.68   & 107.74 & 0.17 \\

    \addlinespace
    \midrule
    \multicolumn{4}{l}{\textbf{Scenario 2}} \\
    \midrule
    \hspace{2ex}FLIE + Target layer graph + D$^{*+}$ (\textit{Baseline}) & 113.755 & 298.94 & 0.28  \\
    \hspace{2ex}GBPlanner2.0                                              & 93.81   & 215.22 & 0.58  \\
    \hspace{2ex}xFLIE (\textbf{Proposed})                                  & 44.66   & 331.40 & 0.195 \\

    \addlinespace
    \midrule
    \multicolumn{4}{l}{\textbf{Scenario 3}} \\
    \midrule
    \hspace{2ex}FLIE + Target layer graph + D$^{*+}$ (\textit{Baseline}) & 118.42  & 506.84 & 0.40 \\
    \hspace{2ex}GBPlanner2.0                                              & 101.40  & 372.48 & 1.37 \\
    \hspace{2ex}xFLIE (\textbf{Proposed})                                  & 52.10   & 573.40 & 0.24 \\
    \bottomrule
    \end{tabular}
  }
\end{table}

Overall, xFLIE consistently maintains the expected behaviour, reducing computational burden upto 60\% during planning compared to the baseline method as the environment scales up. Furthermore, the proposed method covers upto 11.6\% more volume than the baseline method while maintaining upto 40\% lower RMSE during the mission. The lower coverage performance of the baseline method is attributed to the increased response time of the path planner each time the robot chooses to inspect a candidate target node during the later stages of the mission. This delay accumulates over time, negatively affecting overall coverage efficiency under the limited time budget. Moreover, the exploration method achieves a lower inspection volume and consequently a higher RMSE compared to the other methods. However, it maintains approximately 14\%-17\% lower computational overhead compared to the baseline method across the evaluated scenarios. In summary, the results highlight the feasibility of constructing and maintaining an actionable hierarchical scene graph to support autonomous missions in large-scale environments.
 
\section{Lessons Learned and Limitations}\label{sec:lessons}
Developing and deploying the proposed autonomous inspection system in real-world environments revealed certain challenges and design insights that may benefit the community.

Primarily, segmentation quality in the scene processing module proved critical, especially during exploration. Over-segmentation led to inaccurate localization by incorporating background pixels, while our mask erosion fix sometimes produced masks too small to support downstream modules. These errors propagated into semantic localization and graph construction. We tune the erosion parameter to get the best performance based on experimental trials for each semantic class. Because the mask-erosion parameter does not transfer across semantic classes, we set it prior to mission start and keep it fixed for each scenario. Future implementation can look into clustering methods prior to semantic localization such as DBSCAN~\cite{ester1996density} to ensure robust performance.

Furthermore, while optimizing the \textit{Target} layer in the 3DLSG during exploration, we observed that targets detected in non-consecutive frames occasionally bypassed the global optimization step, producing duplicate registrations (Fig 14). These duplicates inflated node centrality and biased evaluation of (2). The root cause was isolated to segmentation inconsistency during fast, unsteady motion, amplified when mask erosion over-shrunk segmented target masks and shifted the centroids estimates beyond the filter threshold $d^T_{check}$. To address this issue during experiments, we chose a more conservative yaw-rate control bounds for a smoother motion. We motivate investigation for an appropriate optimization strategy as part of future studies.

\section{Future Works}\label{sec:future_works}
Building on the utility of 3D scene graphs for inspection planning in unknown environments, our architecture opens the door to two promising extensions. First, the semantic querying capabilities of the 3DLSG can be leveraged for multi-modal task and mission planning, where high-level objectives involving exploration, inspection, or response operations are decomposed into executable actions potentially aided by recent advances in LLM-enabled autonomy~\cite{royce2024enabling,ravichandran2024spine}. Second, the hierarchical structure of the 3DLSG lends itself to multi-agent coordination: one agent may perform exploration and populate higher-level semantic layers, while another conducts detailed inspection and updates lower-level pose and feature layers. This division of labor contributes towards large-scale operations leveraging the inherent modularity of the 3DLSG.
\section{Conclusions}\label{sec:conclusions}
We presented successful field and simulation deployments validating the effectiveness of the xFLIE framework. The proposed method achieves multiple orders of magnitude faster path-planning compared to conventional volumetric approaches in large-scale environments. Furthermore, xFLIE successfully demonstrates incremental construction and exploitation of the hierarchical scene representation to support target selection, path planning, and semantic navigation during autonomous inspection missions. These results collectively highlight the advantages of maintaining a scalable and robot-interpretable hierarchical scene model to support autonomous operation in previously unknown environments.

\bibliography{bib.bib}

\vfill\pagebreak

\end{document}